# Your Ride, Your Rules:

## Psychology and Cognition Enabled Automated Driving Systems

Zhipeng Bao and Qianwen Li*
School of Environmental, Civil, Agricultural, and Mechanical Engineering
University of Georgia
zb28097@uga.edu and cami.li@uga.edu

**Abstract**

Despite rapid advances in autonomous driving technology, current autonomous vehicles (AVs) primarily respond to external traffic conditions and treat humans as passive occupants, lacking mechanisms for active adaptation and collaboration. This limitation constrains their ability to personalize driving behavior to human expectations and hinders effective navigation of ambiguous traffic scenarios that could benefit from leveraging the occupant's advanced cognitive input, resulting in increased delays and potential safety risks. This inadequacy in the long term undermines occupant trust and hinders the widespread adoption of AV technologies. This research is motivated to propose PACE-ADS (Psychology and Cognition Enabled Automated Driving Systems): a human-centered autonomy framework that enables AVs to sense, interpret, and respond to both external traffic conditions and internal occupant states. PACE-ADS is built on an agentic workflow where three foundation model agents collaborate: the Driver Agent interprets the external environment; the Psychologist Agent decodes passive psychological signals (e.g., facial expressions) and active cognitive inputs (e.g., verbal commands); and the Coordinator Agent synthesizes these inputs to generate high-level driving behavior decisions and parameters that enhance responsiveness in ambiguous scenarios and personalize the ride. PACE-ADS is designed to complement, rather than replace, conventional AV modules. It operates at the low-frequency semantic planning layer while delegating low-level, high-frequency control to the vehicle's native systems. It activates selectively in response to changes in the rider's psychological state, cognitive instructions, and vehicle immobilization. It functions as an "add-on" module and easily integrates into existing AV platforms without architectural changes, positioning PACE-ADS as a scalable enhancement for next-generation AVs. We evaluate PACE-ADS in closed-loop simulations using CARLA. Results show that the system dynamically adjusts driving behavior in response to evolving passenger states, enhances riding comfort, and enables safe recovery from operational immobilization through self-reasoning and occupant-in-the-loop guidance.

## I. Introduction

Many automakers have already implemented various automation features in their mass-production vehicles, such as Ford, GM, and BMW. Common features include Adaptive

Cruise Control (ACC), Automatic Lane Changing (ALC), and Lane Keeping Assistance (LKA). The global market for these technologies is projected to grow from $40.25 billion in 2024 to $75.23 billion by 2030, reflecting a strong compound annual growth rate (CAGR) of 13.83% [1]. While lower levels of automation have seen widespread deployment, higher levels (those capable of handling more complex driving tasks with minimal human input) have progressed much more slowly. So far, only Tesla has deployed a Full Self-Driving (FSD) system in mass-produced vehicles. Other companies, such as Baidu and Waymo, have focused on limited deployments through robotaxi services in cities like Beijing, Wuhan, Phoenix, San Francisco, and Los Angeles. The slower advancement of higher-level automation can be attributed to several factors. Beyond technological and regulatory challenges, one of the most significant barriers is low public trust. The underlying reasons for this hesitation are explored in the following discussion.

First, existing automated driving systems (ADS), i.e., higher-level automation, lack the ability to personalize the riding experience and offer very limited transparency in their decision-making processes. Their behavior often appears opaque, leaving most riders uncertain about why the vehicle behaves in a particular way. These systems are primarily designed to optimize safety and efficiency based on external traffic conditions, frequently overlooking the diverse and evolving preferences of individual occupants. When an ADS's driving style significantly deviates from a rider's expectations, acceptance tends to decline [2]. Some progress has been made to address this limitation through the introduction of pre-defined driving styles. For example, Tesla's FSD system allows passengers to choose among "Chill," "Average," and "Assertive" modes. In academic research, human driving data have been used to classify three primary styles: conservative, normal, and aggressive, using methods such as Gaussian Mixture Models [3], K-means clustering [4,5], Gaussian Process models [6], Artificial Neural Networks [2,7], and other unsupervised learning techniques [8,9]. While most studies rely on vehicle trajectory data to infer driving styles, Ling et al. [10] adopt a different approach by assessing driving styles through emotions detected from drivers' electroencephalogram (EEG) signals. They then use the Deep Deterministic Policy Gradient (DDPG) algorithm [11] to adjust control parameters based on the identified driving style.

However, the varying nature of human needs cannot be fully captured by a limited set of pre-defined driving styles. To truly enhance the user experience, ADS must be individualized to account for the diverse and dynamic preferences of occupants. Some studies have attempted to address this challenge by collecting extensive data from individual drivers and using reinforcement learning or imitation learning to train ADS models that replicate specific driving styles [12–17]. These approaches rely on offline training, after which the system is expected to mimic the target driver's behavior. Yet, human psychological states are inherently fluid, shaped by context, emotion, and situational demands. As a result, rider preferences can shift significantly even within the course of a single trip. For instance, a passenger might feel rushed at the beginning of their commute if they believe they are running late but become noticeably more relaxed once they realize they will still arrive in time for their meeting. Offline-trained models often fail to adapt to such real-time changes, resulting in less satisfying experiences and eroding public trust. To address this limitation, research is needed to develop personalized ADS capable of responding to the evolving needs of occupants. A promising direction is the real-time and continuous monitoring of human emotional states. Psychological signals (such as EEG signals and heart rate) and cognitive cues (such as facial expressions and verbal commands) are particularly valuable for this purpose, as they provide immediate feedback in response to external stimuli. Although these indicators have been widely used to detect driver fatigue and distraction [18–21], their potential to inform vehicle control decisions in automated systems remains largely untapped.

Second, ADS struggles in rare or ambiguous traffic scenarios such as construction zones, informal right-of-way negotiations, or unpredictable pedestrian behavior. These situations often lead to the vehicle becoming immobilized [22-25]. For example, in December 2024, a Waymo robotaxi in California became stuck in a roundabout, circling 17 times before engineers had to intervene remotely to resolve the issue [26]. A similar incident occurred in Wuhan, China, where a Baidu robotaxi failed to replan its trajectory in a congested intersection. The vehicle remained stopped for over an hour, causing significant traffic disruption. Local traffic authorities were unable to assist, and control was eventually restored by remote engineers [27]. Although remote intervention provides a potential backup solution, it presents several challenges. It requires continuous monitoring by trained personnel who must be available to respond quickly when the system encounters difficulties. This approach demands substantial human and financial resources, and it also relies on a low-latency connection to ensure timely and effective action. Some companies have developed ADS that allows passengers to take over when the system encounters a failure [28]. For instance, Tesla's shadow mode [29] enables its FSD system to learn from human driving behavior in situations where the vehicle becomes immobilized and to reduce the likelihood of similar failures in the future. However, such solutions rely on the assumption that passengers are capable of taking control. In cases where riders are elderly, have disabilities, or are otherwise unable to take over, the ADS may remain stuck indefinitely. This highlights the need for more inclusive intervention strategies that can complement existing methods. One promising direction is to allow passengers to provide guidance through natural interactions, such as verbal commands, enabling them to assist the system without requiring physical control of the vehicle. Our earlier framework, StuckSolver [30], operationalizes this idea. Results from the Bench2Drive benchmark [31] show that it can autonomously make recovery decisions, interpret passenger instructions, and execute appropriate behavioral plans to restore vehicle mobility.

With their advanced semantic understanding and reasoning capabilities, Large Language Models (LLMs) have emerged as a promising intermediary between occupants and ADS. Recent studies have begun exploring how LLMs can be integrated into ADS frameworks [33-39]. For instance, PromptTrack [40] enhances 3D object tracking and prediction in driving scenes by incorporating cross-modal features through a prompt reasoning branch. DiLu [41] combines LLMs with a reasoning and reflection module, allowing AVs to make decisions based on common-sense knowledge and improve generalization by accumulating driving experience over time. LLMs also show potential in vehicle motion control. For example, Sha et al. [39] use LLMs to interpret driving scenes, make decisions, and convert these decisions into actionable driving commands through a parameter matrix adaptation process. Cui et al. [42] deploy a fine-tuned, lightweight vision-language model (VLM) on an automated vehicle, enabling it to adjust control parameters based on passenger feedback after each trip. While existing studies have made significant strides in integrating LLMs into ADS, most are evaluated using open-loop testing rather than real-time, closed-loop validation. This prevents the ADS from adapting to new inputs during the trip, including rider preferences and situational changes. Without feedback-driven learning, the system cannot refine its decisions dynamically, limiting its effectiveness in real-world, continuously evolving environments.

Motivated by the limitations outlined above, this project introduces PACE-ADS (Psychology and Cognition Enabled Automated Driving Systems), a novel human-centered autonomy framework that empowers vehicles to sense, interpret, and respond to both the external traffic environment and the internal state of the rider. Built on an agentic workflow, PACE-ADS leverages three specialized foundation model agents to collaboratively manage the complex driving tasks, capitalizing on their unparalleled capabilities in multi-modal reasoning and natural language interaction. PACE-ADS comprises: 1) a Psychologist Agent, which interprets passive psychological signals and active cognitive instructions to assess rider psychological state and intent; 2) a Driver Agent, which perceives the surrounding traffic context using sensor data; and 3) a Coordinator Agent, which integrates these inputs to suggest appropriate operation behaviors (e.g., lane changes), determine the value of the operation parameters (e.g., acceleration and headway), and deliver language explanations to the rider upon request. PACE-ADS is not designed to replace traditional real-time control modules, but to complement them by operating at the semantic and behavioral planning level. While real-time modules handle low-latency execution, PACE-ADS contributes in two ways: it monitors human states and proposes adaptive behaviors for personalized riding experience, and it supports vehicle operation recovery during immobilizing scenarios through context-aware reasoning and/or rider-in-the-loop decision support. The system operates in a closed-loop architecture, continuously integrating feedback from the external environment and the rider's internal state to enable safe and progressively adaptive autonomy.

The remainder of this manuscript is organized as follows. Section 2 introduces the PACE-ADS methodology. Section 3 details the simulation experiments conducted to evaluate the model's performance. Section 4 concludes the study and outlines directions for future research.

## II. Psychology and Cognition Enabled Automated Driving Systems

By enabling bidirectional human-machine communication through foundation models (Figure 1), PACE-ADS personalizes the autonomous riding experience and improves vehicle responsiveness in immobilizing scenarios to enhance occupant comfort and trust, ultimately promoting the adoption of automation technologies. PACE-ADS is not designed to replace traditional real-time control modules, but to complement them by operating at the semantic and behavioral planning level. PACE-ADS continuously monitors occupants' passive psychological signals (e.g., EEG, heart rate, facial expressions) and active cognitive instructions (e.g., verbal commands) to infer the occupant's state and intent. In response to such observations, the system dynamically adapts the ego vehicle's driving behavior to personalize the riding experience and enhance user comfort. It also provides behavior explanation using natural language to occupants upon request to enable operation transparency. When the AV becomes immobilized under ambiguous traffic scenarios such as unclear right of way, PACE-ADS leverages the occupant's cognitive instructions (e.g., verbal commands) to execute appropriate recovery solutions. PACE-ADS builds on LLMs given their strengths in semantic understanding, reasoning, and natural language interactions. By implementing an agentic workflow where three specialized agents collaborate to manage the complex autonomous driving, PACE-ADS achieves improved operation interpretability and system robustness. Leveraging system prompts, chain-of-thought (CoT) prompting [43], and function calling tools, each agent is guided to fulfill its designated role, execute assigned subtasks, and produce structured outputs.

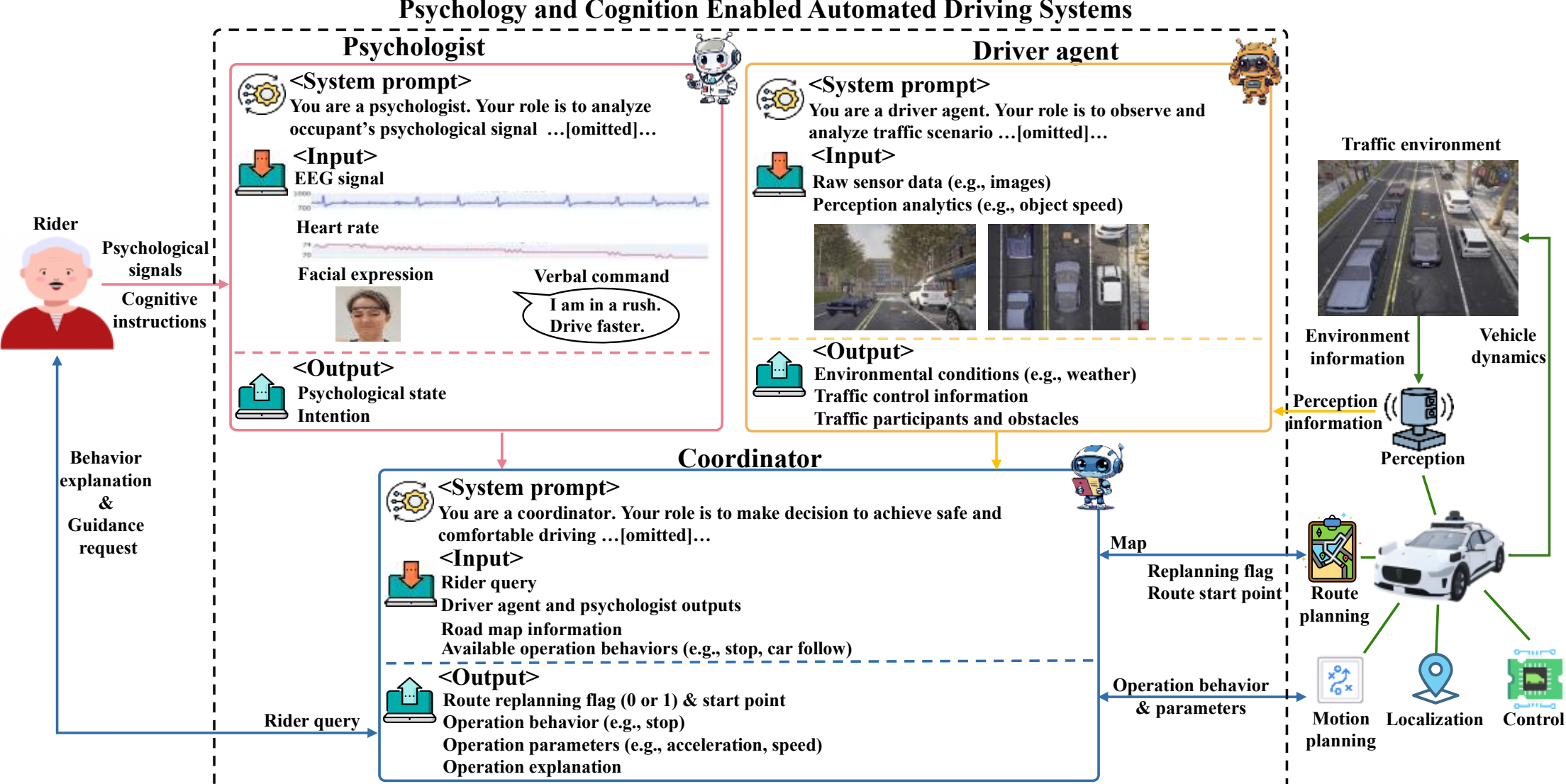

Figure 1. An overview of Psychology and Cognition Enabled Automated Driving Systems.

A key strength of PACE-ADS is its architectural flexibility. The three agents can be configured with either the same or different models, depending on the data modality and task specialization. While this research explores a unified approach by using three LLMs to fulfill all roles, PACE-ADS is open to a hybrid configuration. For example, rule-based models and traditional lightweight machine learning models can be adopted to realize less demanding roles, such as emotion recognition, with a balance of solution quality and efficiency. In addition, PACE-ADS integrates with existing AV systems without requiring structural modifications, as it only receives input from AV modules and returns suggested operations in alignment with the AV module data format. While the AV modules handle high-frequency, low-level control tasks, PACE-ADS focuses on low-frequency, high-level behavioral decisions. It activates selectively in response to rider inputs and vehicle immobilization, addressing non-time-sensitive scenarios through closed-loop feedback that supports adaptive, personalized autonomy without disrupting real-time control.

While PACE-ADS enables adaptive autonomy, all personalization is bounded. Rider-specific adjustments are confined within pre-defined safety envelopes. These envelopes are aligned with traffic laws and social norms to ensure that behavior remains within legally and socially acceptable bounds. In the limiting case, a fully personalized AV may resemble a human-driven vehicle exhibiting user-preferred driving traits, such as assertiveness or caution, yet still operate under the same legal and infrastructural constraints as other road users. This mirrors today's traffic environments, where individuals drive in varied, but lawful, ways. Such personalization improves rider comfort and trust without introducing behavior that is less predictable than conventional human driving.

### A. Existing Autonomous Vehicle Modules

Existing AV frameworks consist of a set of modules that operate in coordination to enable automation. PACE-ADS interfaces with the following modules.

**Perception module.** This module supplies three types of data to the driver agent: RGB images, object-level measurements, and ego vehicle state information. The RGB images include both front-view and bird-eye-view (BEV) perspectives ($I_{front}, I_{bev}$). These images may be captured directly from onboard cameras or generated via sensor fusion and BEV map construction pipelines [44], depending on the specific AV system architecture. The object measurement data $\boldsymbol{O^p}$ includes information for each surrounding object $i$: its class or identifier, relative distance, velocity, and spatial location $O_i^p = \{id, dist, v, loc\}$. The ego vehicle state data includes the velocity and spatial location $Ego = \{v, loc\}$, as detailed in Appendix Table 1.

**Route planning module**. This module is responsible for computing the shortest global route from the starting point to the destination. It begins by retrieving the road topology from the high-definition map and then applies the route planning algorithm, such as the A* algorithm with Euclidean distance. The generated route consists of a series of waypoints that guide the ego vehicle to its destination. We encapsulate the route planning function as an API, enabling the coordinator to invoke it when needed. The API accepts two parameters: a replanning flag $\delta$ and a new route start point $P_{start}$. The data transmitted from the route planning module to the coordinator includes the ego vehicle's current lane, adjacent lanes, and the set of available waypoints located ahead of the vehicle on these lanes, as outlined in the map $M$ of Appendix Table 1.

**Motion planning module.** This module is responsible for generating the vehicle's short-term trajectory along the assigned global route. It manages a library of driving operations $\boldsymbol{A_{lib}}$, such as lane changing, car following, and stopping. The motion planning module continuously evaluates the surrounding environment and the vehicle's current state to make behavior-level decisions. Once a behavior is selected, the corresponding trajectory planning algorithm generates a local trajectory to guide the vehicle's motion. The generated trajectory consists of a series of waypoints, each paired with an expected vehicle speed. Each supported driving operation in $\boldsymbol{A_{lib}}$ is documented with a functional description and a set of configurable parameters as well as default values (Figure 2). The motion planning module operates using its internal decision rules when no external instruction is provided. However, when the coordinator issues an instruction (specifying the desired behavior $a$ and associated parameters $\theta^*$) and safety conditions allow, the module will follow the specified behavior.

```
CAR_FOLLOWING_TOOL = """
Description: this algorithm is based on the Intelligent Driver Model(IDM). It can
control the vehicle to follow the leading vehicle.
Parameters:
- a: the maximum acceleration of the vehicle.
       Threshold: [0.3, 7.0] (m/s^2)
- b: the comfortable deceleration of the vehicle.
       Threshold: [0.5, 7.0] (m/s^2)
- T: the desired time headway to the vehicle ahead.
       Threshold: [0.8, 3.0] (s)
- v_0: the desired speed of the vehicle in free traffic.
       Threshold: [5, 40] (m/s)
Default values: a = 0.75, b = 1.67, T = 1.5, v_0 = 11.
"""

STOPPING_TOOL = """
Description: this algorithm can adjust your car's speed to stop.
Parameters: brake: control the brake of the vehicle. The value range is [0, 1], where 0
means no brake and 1 means full brake.
Default values: brake = 0.5
"""
```

Figure 2. Example documentation of driving operation tools in $\boldsymbol{A_{lib}}$.

**Control Module.** The control module is responsible for executing the planned trajectory by generating low-level control commands, such as steering angle, throttle, and braking. It takes the local trajectory produced by the motion planning module and computes real-time control signals that guide the ego vehicle along the trajectory with stability and precision. The module continuously monitors the vehicle's current state, including position, orientation, and speed, and applies feedback control algorithms (e.g., Proportional-Integral-Derivative algorithm (PID), Model Predictive Control algorithm (MPC)) to minimize deviations from the reference path. By ensuring accurate tracking of the planned motion, the control module plays a critical role in maintaining safety, comfort, and responsiveness in dynamic driving environments.

## B. Driver Agent

Traditional perception modules are typically limited to low-level tasks such as object detection, classification, and tracking. These modules may misinterpret scenes and fail to capture the semantic relationships and interactions among objects [45]. The driver agent addresses this limitation by enabling high-level scene understanding (such as the effects of weather on driving behavior and the interactions among traffic participants) through semantic information extraction and commonsense reasoning $F^{driver}(\cdot)$. It receives front-view and BEV images $(I_{front}, I_{bev})$ from the AV perception module. Guided by a system prompt $S^{driver}$ and CoT prompting $C^{driver}$, the driver agent follows a step-by-step reasoning process (see Figure 3) that includes extracting environmental conditions such as weather and road type, identifying traffic control elements like traffic signal state, traffic sign content, and work zones, and analyzing traffic participants and obstacles by determining their types, relative positions (e.g., on the left), and intentions.

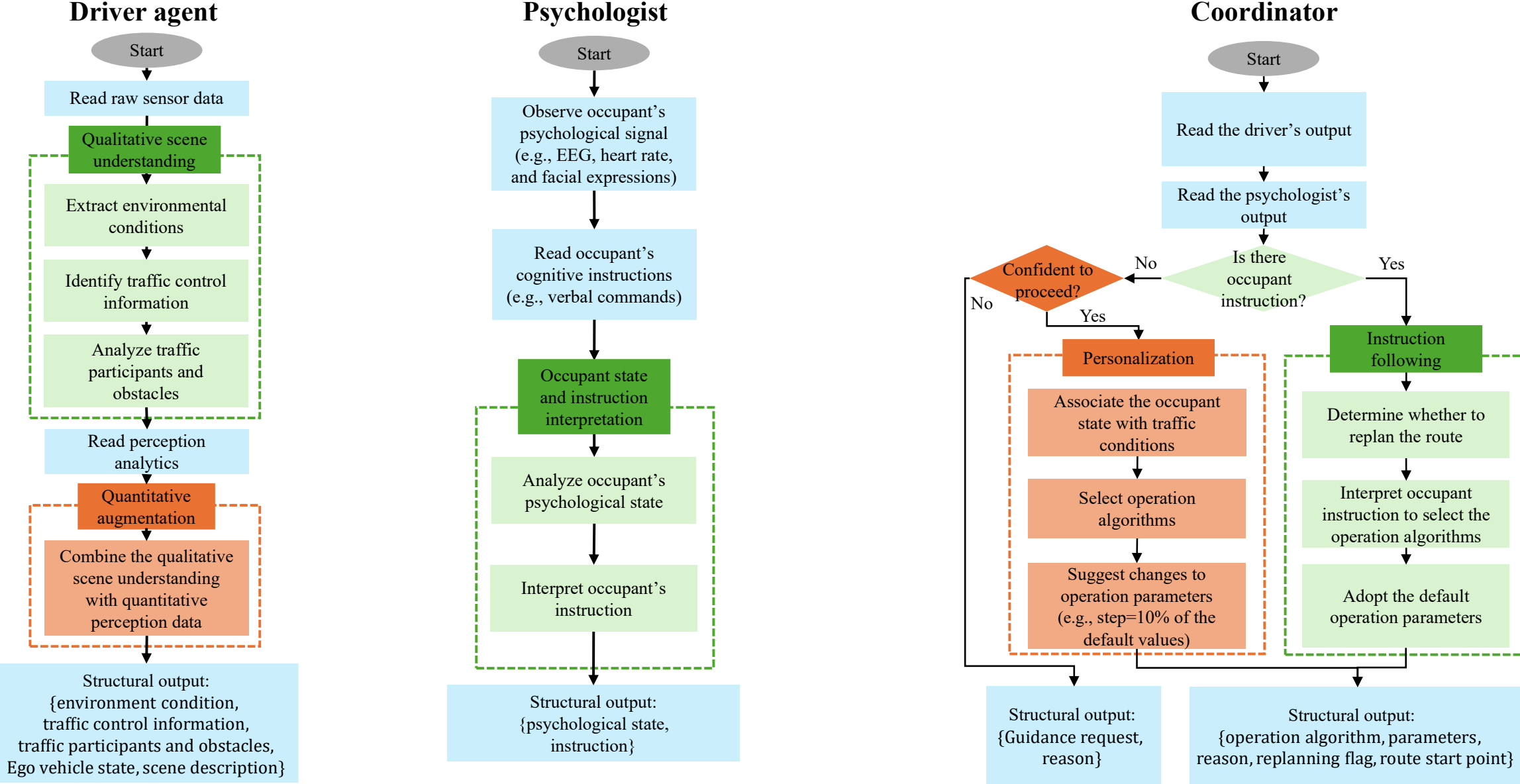

Figure 3. A schematic illustration of the reasoning processes of the three agents.

The driver agent associates the resulting high-level insights with corresponding perception analytics ( $\boldsymbol{O^p}$ ) from the perception module and enriches them with quantitative attributes such as distance, velocity, and lane ID to facilitate context-aware decision-making. The resulting structured output $\boldsymbol{Y}$ includes environmental conditions $E$, traffic control information $T$, traffic participants and obstacles information $\boldsymbol{O}$, ego vehicle state $Ego$, and a scene description $D$, as shown in the equation below. This comprehensive scene understanding is passed to the coordinator.

$$\boldsymbol{Y} = \{E, T, \boldsymbol{O}, Ego, D\} = F^{driver}\left(I_{front}, I_{bev}, \boldsymbol{O^p}; S^{driver}, C^{driver}\right).$$

A detailed description of the input and output components is provided in Appendix Table 1.

### C. Psychologist

The psychologist agent analyzes the occupant's passive psychological signals and active cognitive instructions by processing multimodal inputs, including EEG signals $EEG$, heart rate $HR$, facial expression images $I_{face}$, and verbal commands $Ver$. Guided by the system prompt $S^{psy}$ and CoT prompting $C^{psy}$, the psychologist agent follows a step-by-step reasoning process ($F^{psy}$) to interpret these signals and synthesize a comprehensive understanding of the occupant's state and needs (see Figure 3). The resulting output is denoted by $R = \{state,\ instr\}$, where $state$ is the emotional state of the rider and $instr$ is the explicit rider commands. $R$ is delivered to the coordinator for human-centric decision-making.

$$R = \{state, instr\} = F^{psy}\left(EEG, HR, I_{face}, Ver; S^{psy}, C^{psy}\right)$$

A detailed description of the input and output components is provided in Appendix Table 1.

### D. Coordinator

The coordinator is guided by the system prompt $S^{coord}$ and CoT prompting $C^{coord}$, which defines its responsibilities and outlines the sequential procedures for decision-making. It prioritizes driving safety first, followed by personalization to accommodate the occupant's preferences. The coordinator reads the augmented traffic semantic information $\boldsymbol{Y}$ provided by the driver agent and references the psychologist's occupant report $R$. If no explicit occupant instruction is available, the coordinator interprets the occupant's state to select an appropriate operation behavior ($a$) from the vehicle motion planning module and sets the corresponding operation parameters ($\theta^*$). When faced with uncertainty, the coordinator will actively seek guidance from the occupant, representing an enhancement over our previous work, StuckSolver [30]. When an occupant instruction is available, the coordinator enters instruction-following mode. It first determines whether route replanning is necessary. Then, using the instruction along with the augmented traffic semantic information $\boldsymbol{Y}$, the coordinator selects an appropriate operation behavior. If the instruction does not specify parameter values, the coordinator applies the default parameters associated with the selected behavior. Otherwise, the parameters are inferred from the instruction. If route replanning is needed, the coordinator generates a replanning flag ($\delta$, binary) and specifies a new route start point ($P_{start}$) based on the map $M$, then passes both to the vehicle route planning module. The overall decision-making workflow of the coordinator is illustrated in Figure 3.

$$N = \{a, \theta^*, r, \delta, P_{start}\} = F^{coord}(\boldsymbol{Y}, R, M; S^{coord}, C^{coord}, \boldsymbol{A_{lib}})$$

Here, $N$ represents the coordinator's structured output, $F^{coord}$ denotes the coordinator's reasoning process, $a \in \boldsymbol{A_{lib}}$ is the selected operation behavior from the behavior library $\boldsymbol{A_{lib}}$, offered by the vehicle motion planning module, and $r$ is the explanation for the selected decision. A detailed description of the input and output components is provided in Appendix Table 1.

The coordinator also passively detects potential vehicle immobilization by observing the outputs of the route planning and motion planning modules. Immobilization is suspected when the route planner indicates continued progress is expected, yet the motion planner shows repeated stops, failed trajectory generation, or prolonged stationary behavior without external justification (e.g., traffic signals or destination reached). When such inconsistencies persist, the coordinator flags a possible immobilization event and activates the driver agent to analyze the surrounding environment and guide further action.

## III. Experiment

We design a set of scenarios to evaluate whether PACE-ADS can successfully adapt its driving behavior based on the occupant's state to provide a comfortable and personalized riding experience. Additionally, we evaluate the system's ability to assist an immobilized AV by following verbal instructions from the occupant.

### A. System Implementation Details

All three agents in PACE-ADS are built on GPT-4o-2024-08-06 [46], which is equipped with multimodal processing capabilities, enabling the understanding and analysis of images and text. The model also supports structured output, facilitating seamless interaction and integration with existing AV modules. Each agent is encapsulated as a ROS2 (Humble version) node [47] and connected to the CARLA (0.9.15 version) server through the carla-ros-bridge, forming a closed-loop simulation platform for system validation and testing. All simulations are conducted on a workstation running Ubuntu 22.04, equipped with an AMD Ryzen Threadripper 7960X (24 cores) CPU and two NVIDIA RTX 5000 Ada GPUs.

We build upon the Behavior Agent provided by the CARLA [48], which integrates key components including the perception module, route planning module, and control module. Specifically, we enhance the perception module by equipping it with multiple sensors, including a front-view RGB camera, a top-down RGB camera, an Inertial Measurement Unit, a Global Navigation Satellite System, and a speedometer. These sensors enable the collection of comprehensive ego vehicle status and surrounding environmental data within the CARLA simulation environment. The route planning module in Behavior Agent generates a global route between a specified starting point and a target destination using the A* algorithm [49]. The resulting global route is represented as an ordered sequence of waypoints along the road network, providing high-level navigation guidance for downstream modules. The control module unifies motion planning and low-level control. It interprets perception inputs and vehicle state information to determine appropriate driving behaviors and executes the corresponding control commands. Car-following and cruising rely on the Intelligent Driver Model (IDM) [50], with default parameters that include a desired velocity of 11 m/s, a safe time headway of 1.5 s, a maximum acceleration of 0.75 $m/s^2$, a comfortable deceleration of 1.67 $m/s^2$, an acceleration exponent of 4, and a minimum distance of 1.5 m.

Stopping behavior is controlled by directly adjusting the ego vehicle's brake pedal input in CARLA, where the brake intensity ranges from 0 to 1, and the default brake value is set to 0.5. Lane changing decision-making is governed by a gap–TTC decision model, in which the model evaluates both the lead and rear gaps ($dist_f$, $dist_r$) as well as the corresponding time-to-collision values ($TTC_f$, $TTC_r$) in the target lane. The default values are set to $dist_f = 5\,\text{m}$, $dist_r = 18\,\text{m}$, $TTC_f = 2.0\,\text{s}$, and $TTC_r = 2.0\,\text{s}$. A lane change is executed once all these conditions are satisfied. Lane changing motion is governed by using separate PID controllers for both lateral and longitudinal directions. For longitudinal control, the proportional, integral, and derivative gains are set to $K_P = 1.0$, $K_I = 0.05$, and $K_D = 0$. For lateral control, the corresponding gains are $K_P = 1.95$, $K_I = 0.05$, and $K_D = 0.2$.

**B. Occupant Data**

We use the publicly available Emognition dataset [51] to provide the passive psychological signals required by the psychologist agent. The dataset includes EEG signals, heart rate data, and facial expression images from 43 participants (see Figure 4), along with annotated emotions such as anger, amusement, sadness, anxiousness, liking, relaxation, and fear. To align with prior studies on drivers' emotional states, we retain the categories relaxed and anxious, and remap anger to impatient. This adjustment reflects the behavioral and physiological similarity between anger and impatience in driving contexts, where frustration and a desire for quicker responses are more accurately described as impatience.

To further examine the responsiveness of PACE-ADS to different levels of emotional intensity, we introduce two additional categories: very impatient and very anxious. Specifically, within the impatient and anxious categories derived from the Emognition dataset, we further distinguish intensity levels based on participants' self-reported emotion values (1 = not at all, 5 = extremely), as offered by the Emognition dataset. Samples with reported values greater than or equal to 4 are labeled as "very impatient" or "very anxious," while those with values below 4 remain labeled as "impatient" or "anxious." Importantly, our intent is not to establish a new taxonomy of passenger or driver types. Instead, our framework assumes that psychological states are fluid and continuous, and cannot be fully captured by a small set of fixed categories. The five emotional states we employ should therefore be understood not as an exhaustive classification, but as interpretive constructions that help us evaluate and communicate how PACE-ADS adapts its behavior in response to changing psychological inputs.

While EEG signals in text, heart rate data in text, and facial expression images are directly fed to the psychologist agent, verbal inputs from the occupant are first transcribed using Whisper [52], a speech-to-text model.

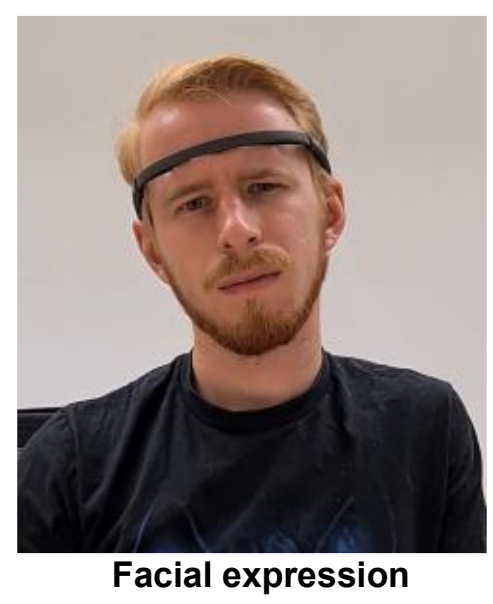

Facial expression

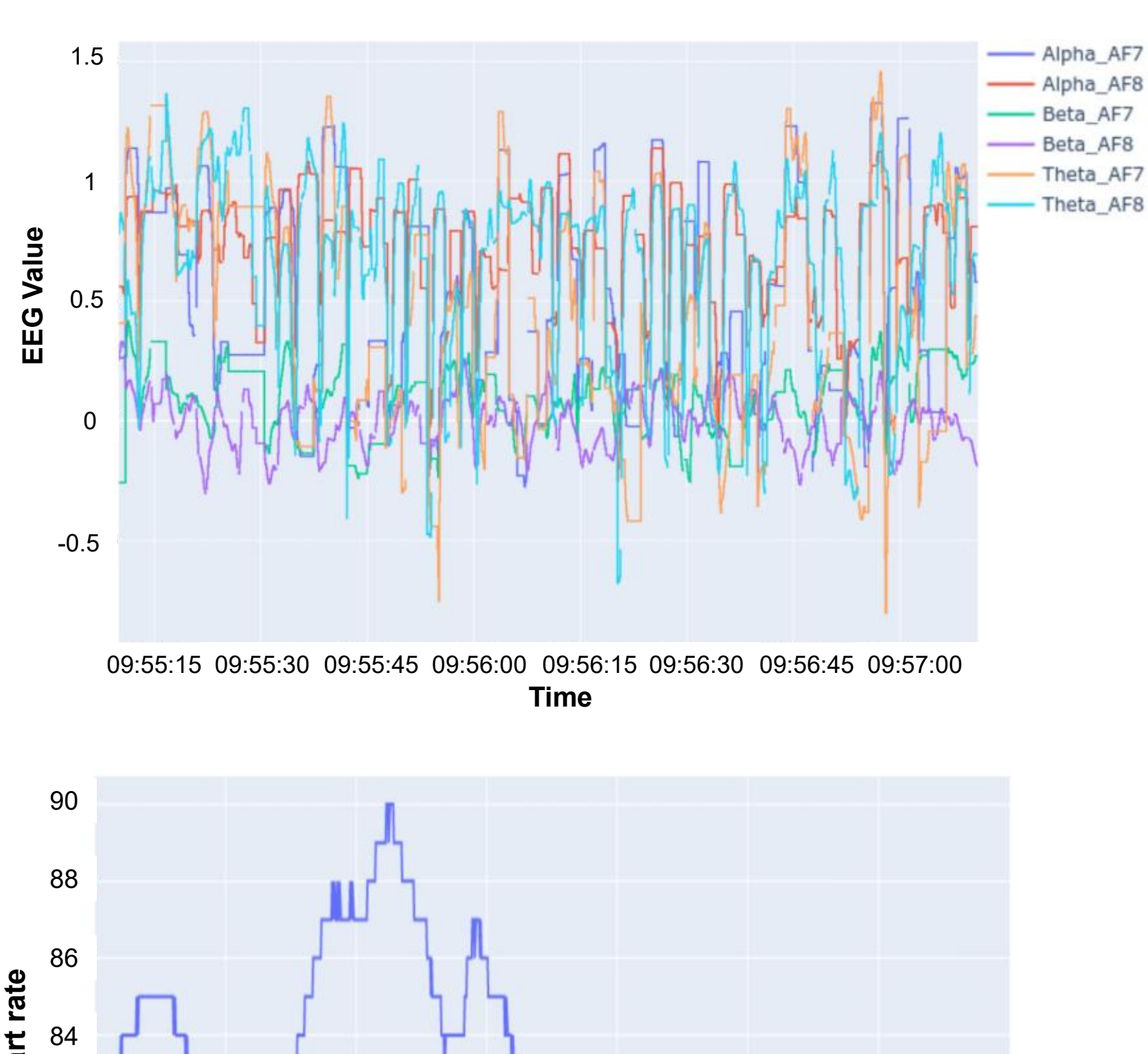

Figure 4. Multimodal psychological signals of the same individual.

## C. Personalization Tests

To evaluate the ability of PACE-ADS to personalize autonomous rides, we design four representative scenarios that capture diverse rider preferences: traffic signal interaction, pedestrian interaction, work zone navigation, and car-following. In each scenario, the occupant's psychological state is simulated by combining multimodal signals representing very impatient, impatient, relaxed, anxious, and very anxious states from the same individual across multiple test runs. Some runs reflect natural fluctuations in passenger state, while others are intentionally exaggerated to rigorously assess system adaptability. This setup enables evaluation of whether PACE-ADS can accurately perceive and respond to a broad spectrum of psychological cues.

**Traffic signal interaction.** The ego vehicle starts at 40 km/h, approaching a red traffic light ahead. Once the vehicle comes to a complete stop, the signal immediately turns green, prompting the vehicle to resume driving. To evaluate the system’s ability to balance occupant preferences with regulation requirements, a maximum speed of 60 km/h is imposed. At the beginning of each run, we inject psychological signals representing different emotional states to simulate how an occupant might respond to the AV’s behavior. When the AV’s behavior aligns with their preferences, signals corresponding to a relaxed state are injected. Otherwise, impatient and/or anxious signals are injected to reflect the rider’s discomfort.

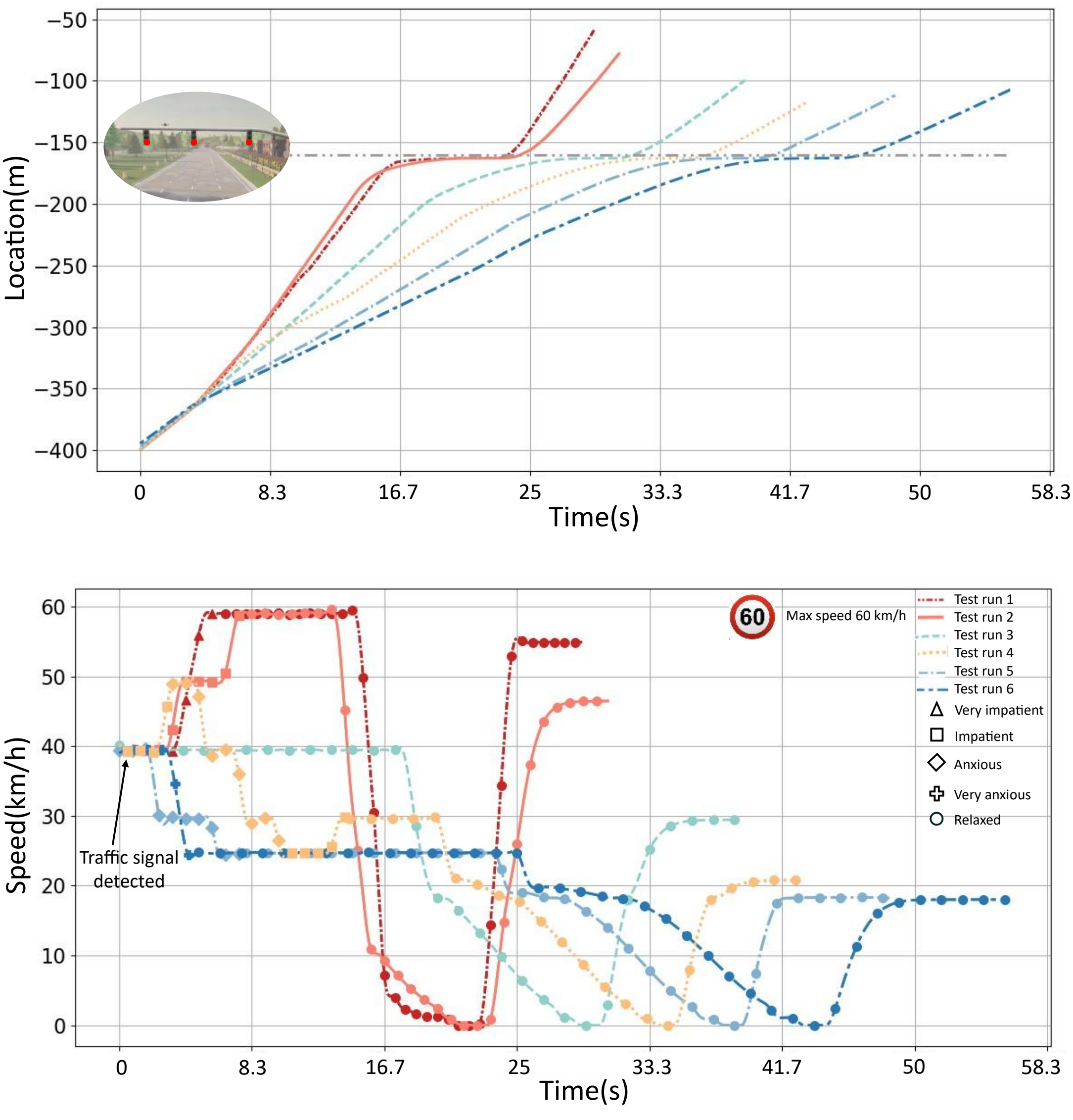

Figure 5. Traffic signal interaction experiment results.

The experimental results, illustrated in Figure 5, show that across different test runs, PACE-ADS successfully recognizes the occupant’s psychological state and responds with appropriate adjustments to driving behavior. The system modifies cruising speed, deceleration timing, deceleration rate, and acceleration in a manner tailored to the occupant's state, enhancing comfort and delivering a personalized riding experience.

Test run 3 serves as a benchmark, with relaxed signals injected throughout the simulation run to represent an occupant who remains consistently satisfied with the AV's behavior. In Test run 1, signals corresponding to a very impatient state are injected from the beginning to reflect dissatisfaction with the initial speed. Once the vehicle accelerates significantly (in this case, toward the maximum permitted speed of 60 km/h), we inject relaxed signals to simulate restored comfort. Under the very impatient condition, the system observes a stronger emotional response and immediately accelerates to the maximum allowed speed of 60 km/h to promptly address the discomfort. During the red light interaction phase, PACE-ADS tailors its deceleration behavior based on previously observed emotional states. It infers a strong preference for maintaining speed and selects the latest possible deceleration point, applying a relatively high deceleration rate to ensure safety while aligning with the occupant's preferences. In the restart phase, the system uses memory of prior emotional states to inform acceleration. It applies higher acceleration rates of 5.4 $m/s^2$ and resumes at elevated cruising speeds of 55 km/h. Test run 2 follows the same structure as Test run 1, but starts with impatient signals instead. Once the vehicle reaches 60 km/h, relaxed signals are injected. In the impatient condition, the system determines that the vehicle is still a considerable distance from the red light and infers that the occupant's impatience stems from dissatisfaction with the current speed. In response, PACE-ADS gradually increases the cruising speed to 50 km/h. When it detects continued impatience, it further raises the speed to 60 km/h. During the red light interaction phase, the system follows a similar deceleration pattern as Test run 1, but with a smaller deceleration. In the restart phase, it applies the acceleration rate of 4.2 $m/s^2$ and resumes at elevated cruising speeds of 45 km/h. In both Test runs 1 and 2, the vehicle remains within the posted speed limit, ensuring regulatory compliance while providing a responsive and personalized experience.

Test runs 5 and 6 follow a similar pattern to Test runs 1 and 2, using anxious and very anxious signals, respectively. For the anxious and very anxious conditions, the system adopts a comparable strategy in the opposite direction. It gradually reduces the cruising speed to address the occupant's heightened discomfort and enhance their sense of safety. During the red light interaction phase, the system chooses an earlier and more gradual deceleration to enhance comfort and reduce perceived stress. In the restart phase, it selects a low acceleration rate of 1 $m/s^2$ and a conservative cruising speed of 20 km/h, aligning with their comfort needs. In Test run 4, we simulate a more dynamic occupant state by injecting a sequence of emotional signals. The test begins with impatient signals. After the vehicle accelerates, anxious signals are injected to reflect growing discomfort. When the speed drops below 30 km/h, impatient signals are reintroduced to indicate a desire for faster travel. Once the speed increases again, relaxed signals are injected to reflect renewed satisfaction. Additionally, where the occupant's psychological state changes frequently, PACE-ADS promptly recognizes each shift and adapts its driving behavior accordingly. This responsiveness helps maintain occupant comfort by continuously aligning vehicle behavior with the occupant's evolving preferences.

Each simulation scenario was repeated 30 times, varying the initial distance between the vehicle and the traffic light from 200 m to 300 m. The average results are summarized in Table 1, including the adjusted velocity ($v_a$) after emotional adaptation, the cruising velocity ($v_c$) after restart, the deceleration rate ($b$) when approaching the red light, and the acceleration rate ($a$) during restart. When the occupant is very impatient, the vehicle exhibits the highest speeds ($v_a = 59.6$ km/h, $v_c = 55.7$ km/h), while under the very anxious condition, both values are the lowest ($v_a = 26.3$ km/h, $v_c = 19.3$ km/h). Moreover, the very impatient state leads to the most aggressive acceleration ($a = 4.23$ m/s²) and deceleration ($b = 5.51$ m/s²) behaviors, whereas the very anxious state results in the smoothest and most conservative maneuvers ($a = 1.78$m/s², $b = 0.42$ m/s²).

Table 1. Personalized evaluation results in the traffic signal interaction scenario (averaged over 30 runs).

| Occupant state | Metrics | | | |
|---|---|---|---|---|
| | $v_a(km/h)$ | $v_c(km/h)$ | $b(m/s^2)$ | $a(m/s^2)$ |
| Very impatient | 59.6 | 55.7 | 5.51 | 4.23 |
| Impatient | 55.3 | 50.1 | 4.63 | 3.59 |
| Normal | 40.0 | 32.3 | 1.02 | 2.08 |
| Anxious | 27.1 | 19.6 | 0.49 | 1.86 |
| Very anxious | 26.3 | 19.3 | 0.42 | 1.78 |

**Pedestrian interaction.** The ego vehicle cruises at 40 km/h. A pedestrian stands ahead on the left side of the road, intending to cross the road. The pedestrian begins to cross after the vehicle comes to a complete stop. Once the pedestrian clears the conflict zone, the ego vehicle resumes driving. Five test runs are conducted, with each run initialized using a distinct psychological signal to emulate the occupant's emotional state. The test conditions are as follows: Test run 1 (very impatient), Test run 2 (impatient), Test run 3 (relaxed), Test run 4 (anxious), and Test run 5 (very anxious), as shown in Figure 6.

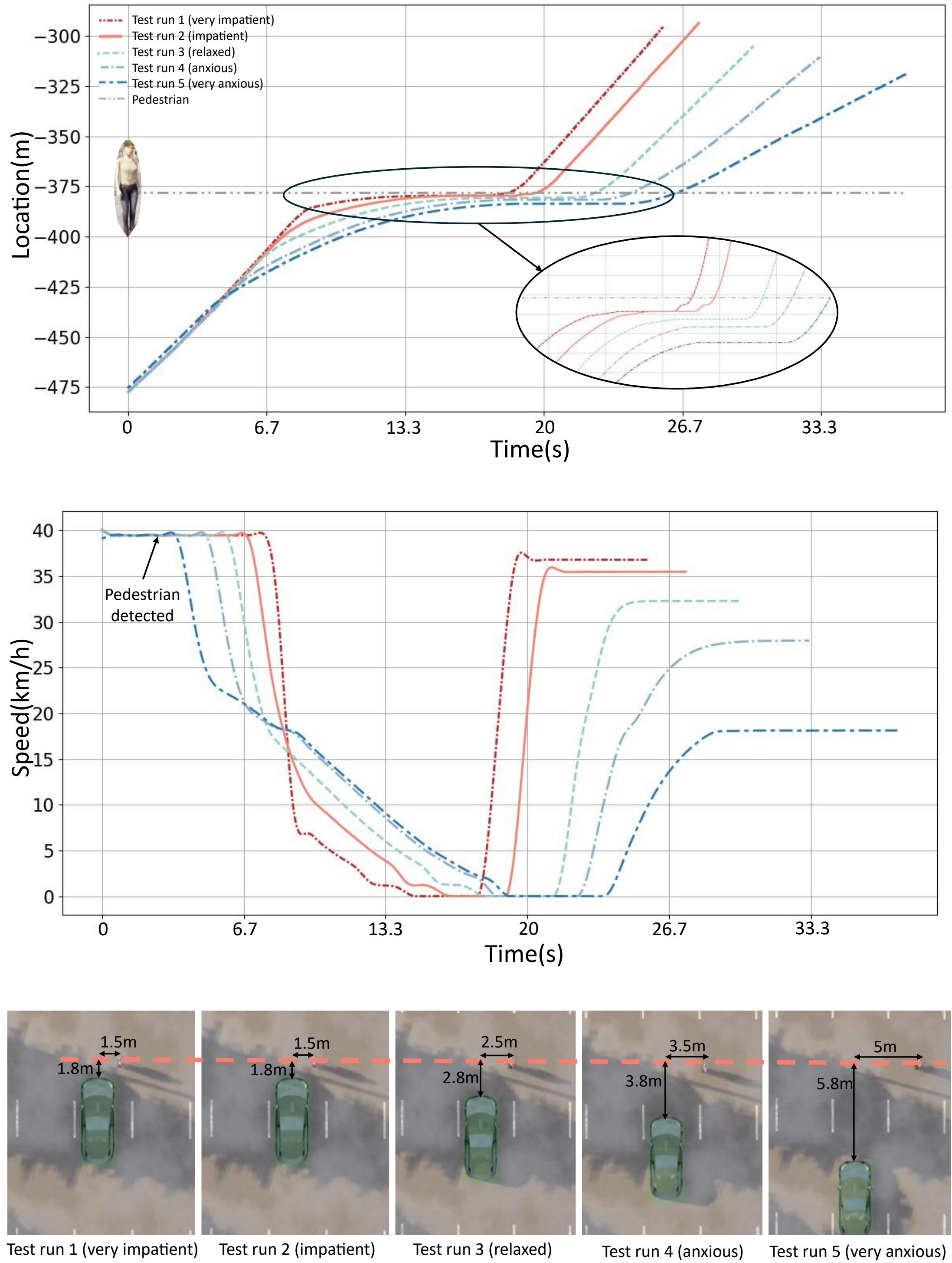

Figure 6. Pedestrian interaction experiment results.

Upon pedestrian detection, the timing of deceleration varies based on the occupant's emotional state. The system initiates stopping earlier for more anxious occupants and later for impatient ones. The sequence of deceleration time, from earliest to latest, is very anxious > anxious > relaxed > impatient > very impatient. Deceleration rates reflect the level of aggressiveness associated with each emotional state. Impatient occupants prompt sharper deceleration, while anxious occupants lead to more gradual braking. The stopping distance and the restart timing, acceleration profile, and cruising speed after the pedestrian clears the path both vary with emotional state. In Test runs 1 and 2 (very impatient and impatient), the vehicle stops closest to the pedestrian at 1.8 meters and resumes motion when the pedestrian is just 1.5 meters away in the lateral direction. The vehicle exhibits a rapid acceleration phase, reaching 35 and 37 km/h with a 3.67 m/s² acceleration rate. In contrast, in Test runs 4 and 5 (anxious and very anxious), the vehicle maintains a larger buffer of 3.8 and 5.8 meters, indicating that PACE-ADS adopts a more conservative stopping strategy under higher anxiety. During restarting, the vehicle waits until the pedestrian moves more than 3.5 and 5 meters away. It accelerates gradually, reaching lower cruising speeds of 28 and 18 km/h with 1.1 and 0.74 m/s² acceleration rates, respectively.

Each simulation scenario was repeated 30 times, varying the initial distance between the vehicle and the pedestrian from 100 m to 200 m. The average results are summarized in Table 2, including the vehicle–pedestrian distance at the onset of deceleration ($\Delta d$), which measures how far the vehicle is from the pedestrian when it begins to slow down; the longitudinal distance to the pedestrian ($d_{lon}$) when the vehicle is stationary; and the lateral distance to the pedestrian ($d_{lat}$) when the vehicle resumes motion. To ensure pedestrian safety, the agents were instructed to maintain a minimum longitudinal and lateral distance of 1.5 m from pedestrians. PACE-ADS effectively accommodates passengers' personalized needs while consistently keeping both distances above the 1.5 m safety threshold, even under the most aggressive (very impatient) passenger state during stopping and restarting ($d_{lon} = 1.67\ m, d_{lat} = 1.62$ m). Under more cautious emotional states, the system exhibits noticeably larger clearances with $d_{lon} = 4.14$m and $d_{lat} = 3.53$ m for the anxious state, and $d_{lon} = 6.04$ m and $d_{lat} = 5.25$ m for the very anxious state.

Table 2. Personalized evaluation results in the pedestrian interaction scenario (averaged over 30 runs, with cruising speed limited to 40 km/h).

| Occupant state | Metrics | | |
|---|---|---|---|
| | $\Delta d(m)$ | $d_{lon}(m)$ | $d_{lat}(m)$ |
| Very impatient | 24.3 | 1.67 | 1.62 |
| Impatient | 35.4 | 1.72 | 1.68 |
| Normal | 46.6 | 2.93 | 2.47 |
| Anxious | 59.9 | 4.14 | 3.53 |
| Very anxious | 78.8 | 6.04 | 5.25 |

**Work zone navigation.** A work zone is placed ahead in the AV's current lane, prompting the vehicle to initiate a lane change to the left. Multiple vehicles are randomly positioned in the adjacent left lane, with varying gaps between them. At the start of the scenario, both the AV and surrounding vehicles travel at a constant speed of 40 km/h. We conduct five test runs, each with injected psychological signals representing the occupant's emotional state. The test conditions are as follows: Test run 1 (very impatient), Test run 2 (impatient), Test run 3 (relaxed), Test run 4 (anxious), and Test run 5 (very anxious).

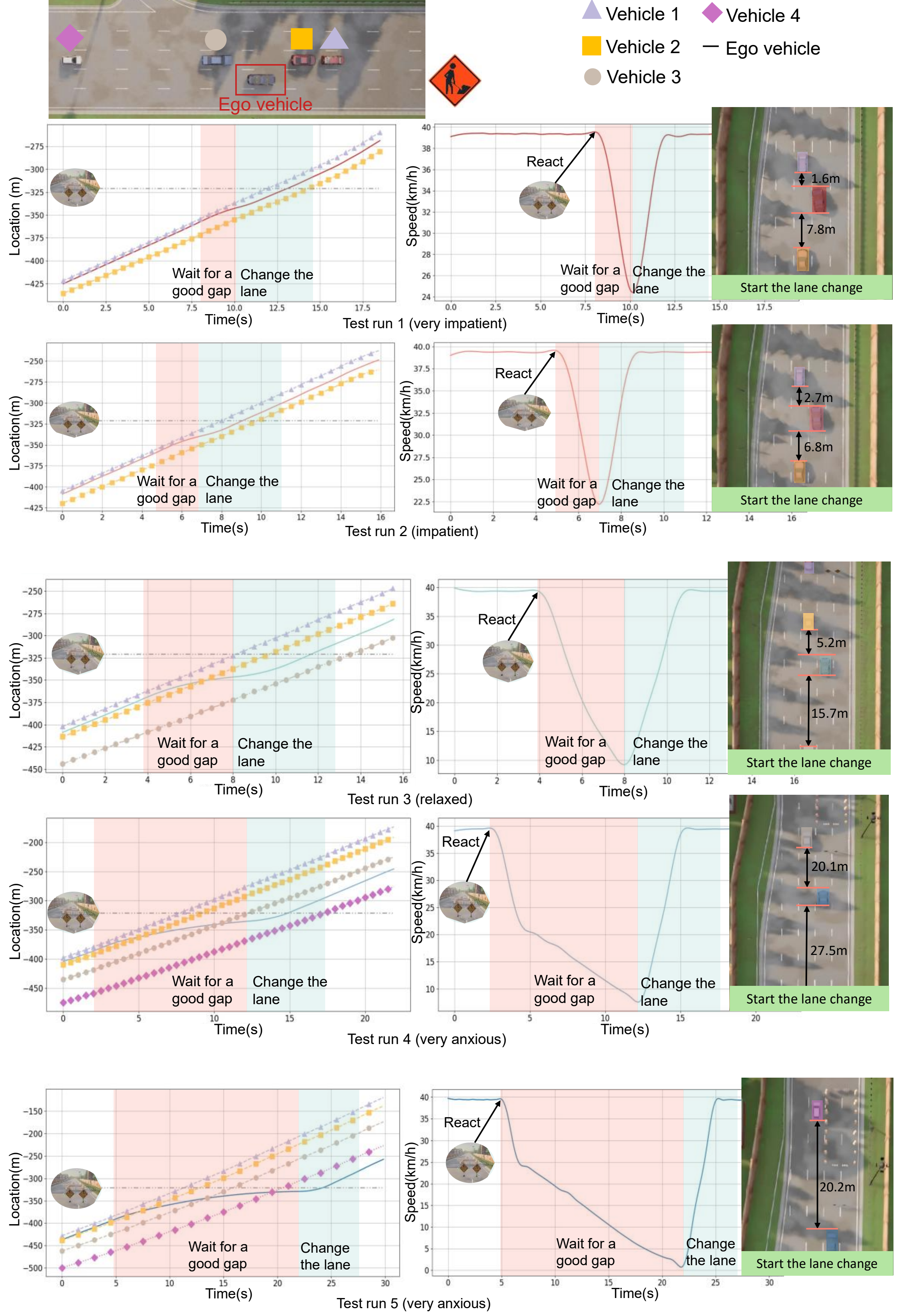

Figure 7. Work zone navigation experiment results.

Figure 7 highlights how PACE-ADS adapts its lane-changing behavior in response to the occupant's psychological state. When the occupant is very impatient (Test run 1), the AV initiates the lane change earliest, performing a quick and assertive maneuver into a relatively small front gap (1.6 m) while maintaining higher speeds throughout. In the impatient condition (Test run 2), the behavior remains proactive but slightly more restrained, with a lane-change speed of 22.5 km/h and a front gap of 2.7 meters. In both runs, the AV accepts the first available gap in the left lane between Vehicle 1 and Vehicle 2. In contrast, for the relaxed condition (Test run 3), the AV selects a lower lane-change speed of 10 km/h and a slightly longer front gap (5.2 m) between Vehicle 2 and Vehicle 3, resulting in a more comfortable maneuver. In the anxious condition (Test run 4), the vehicle begins decelerating early and waits for a significantly larger gap of 20.1 meters before Vehicle 4. The lane-change speed is reduced to below 10 km/h. In the very anxious case (Test run 5), the AV decelerates to nearly a full stop, accompanied by earlier and smoother deceleration profiles. It delays the lane change until all four vehicles have passed and no following vehicle is present in the target lane, ensuring the largest possible gap of 20.2 m before initiating the maneuver.

Each test run was repeated 30 times using different random seeds, with vehicles in the adjacent lane generated at random spacings and with varying numbers. The average results are summarized in Table 3, including the distances to the rear and front vehicles ($dist_r$ and $dist_f$) and the time-to-collision value ($TTC_r$) for the following vehicle in the target lane when the ego vehicle initiates the lane change. As passenger anxiety level increases, both $dist_r$ and $dist_f$ grow significantly, while the $TTC_r$ also increases from 1.77 s to 2.74 s, indicating a more cautious decision-making process. In particular, when the passenger is very anxious, the AV consistently delays its maneuver until no vehicles are present in the adjacent lane behind.

Table 3. Personalized evaluation results in the work zone navigation scenario (averaged over 30 runs, with cruising speed limited to 40 km/h).

| Occupant state | Metrics | | |
|---|---|---|---|
| | $dist_r(m)$ | $dist_f(m)$ | $TTC_r(s)$ |
| Very impatient | 6.4 | 1.52 | 1.77 |
| Impatient | 6.6 | 2.65 | 1.82 |
| Normal | 18.8 | 5.47 | 2.03 |
| Anxious | 28.8 | 20.4 | 2.74 |
| Very anxious | \ | 21.2 | \ |

**Car following.** The lead vehicle follows a predefined speed profile, which simulates traffic oscillation dynamics: it initially travels at 40 km/h, then accelerates to 60 km/h, and finally decelerates to 30 km/h. The initial distance between the ego vehicle and the lead vehicle is set to 10 meters.

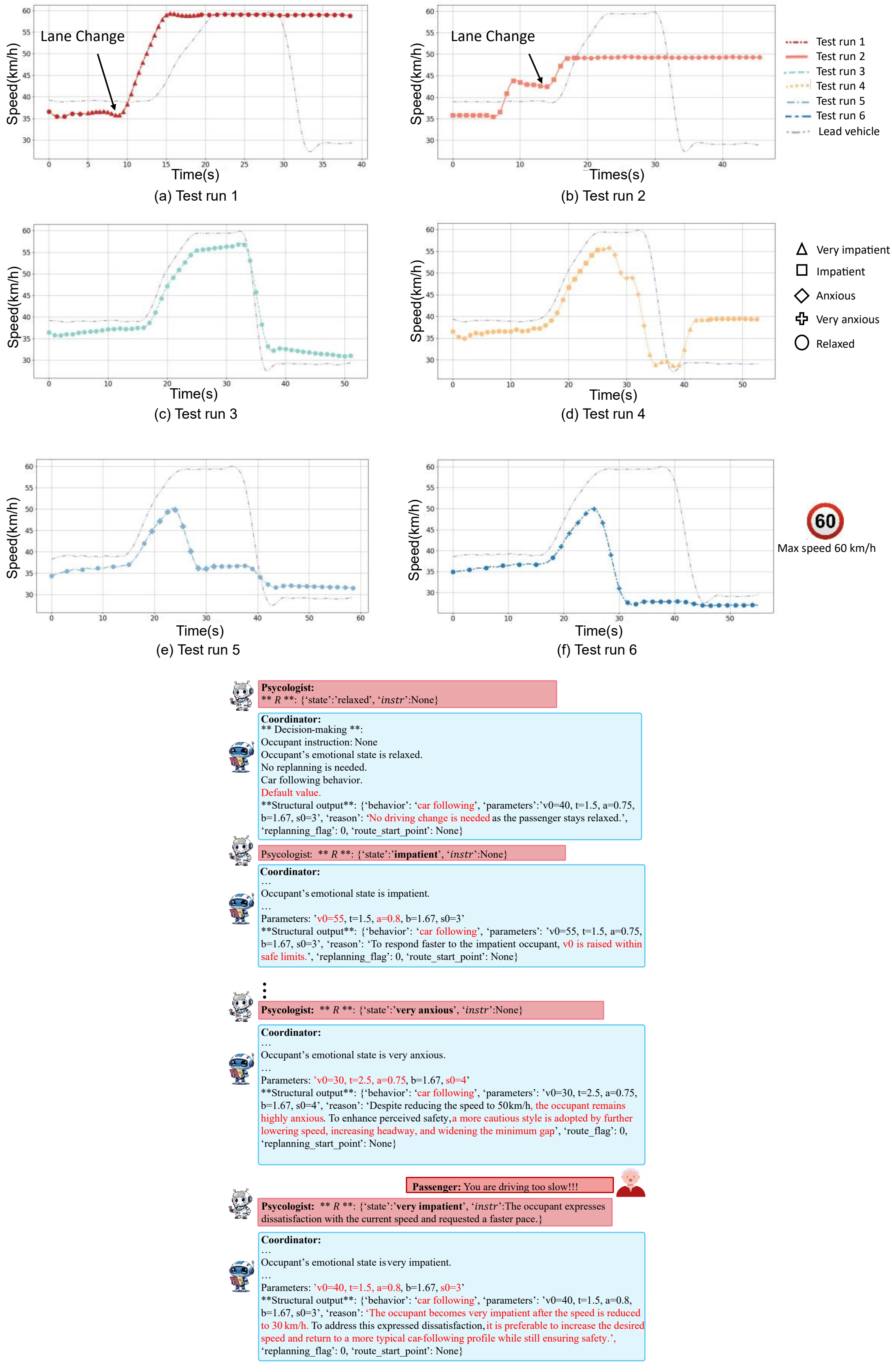

(a) Test run 1 (b) Test run 2

(c) Test run 3 (d) Test run 4

(e) Test run 5 (f) Test run 6

(g) Agent collaboration process in Test run 4

Figure 8. Car following experiment results.

Test run 3 (as shown in Fig. 8(c)), in which the occupant remains relaxed throughout the simulation, PACE-ADS maintains its car following behavior, using the default parameters of the IDM to continuously track the lead vehicle. Test runs 1 and 2 are injected with signals corresponding to very impatient and impatient states, respectively, reflecting occupant dissatisfaction with the initial AV behavior. Once the system increases speed noticeably, signals representing a relaxed state are injected to simulate restored comfort. Fig. 8(a) shows that when the occupant is detected as very impatient in the scenario, the system switches from car following to a lane changing maneuver to promptly address the perceived urgency. It selects an adjacent lane with no surrounding traffic in proximity and accelerates rapidly to the maximum safe speed of 60 km/h. Once the occupant's state returns to a relaxed state, the system continues to maintain the 60 km/h cruising speed. In Fig. 8(b), when the occupant is in an impatient state, PACE-ADS first attempts to increase the following speed to 45 km/h while monitoring for further changes in the occupant's state. When it detects continued dissatisfaction, the system initiates a lane change. After successfully changing lanes, the vehicle accelerates to 50 km/h. Upon observing that the occupant becomes relaxed, the system stabilizes at that cruising speed.

Test runs 5 and 6 follow the same pattern but represent anxious and very anxious emotional states. In Figs. 8(e) and 8(f), the occupants become anxious and very anxious, respectively, in response to the vehicle accelerating while following the lead vehicle. In both cases, PACE-ADS adjusts its behavior by reducing speed. In the anxious case (Fig. 8(e)), the system no longer follows the lead vehicle closely and performs a gradual deceleration—first returning to the initial following speed of 35 km/h, then further reducing to 25 km/h—demonstrating a progressive response aligned with the occupant's emotional state. When the lead vehicle decelerates sharply to 30 km/h and the AV approaches at a higher speed, it further decelerates under IDM control to maintain a safe following distance. For the very anxious occupant (Fig. 8(f)), the system reacts more decisively, immediately reducing speed to 25 km/h. Once the occupant transitions to a relaxed state, the vehicle maintains the adjusted speed. Similarly, when the AV approaches the lead vehicle again, the IDM controller further reduces the speed to ensure a safe following distance.

In Test run 4, as shown in Fig. 8(d), the occupant exhibits frequent changes in emotional state (not necessarily realistic), and PACE-ADS continuously adjusts the vehicle's behavior to align with these dynamic shifts. Specifically, when the occupant becomes impatient, the system increases speed, whereas it reduces the speed when the occupant becomes very anxious. Notably, when the AV decelerates to 30 km/h and the occupant becomes very impatient, the system does not initiate an immediate lane change, as observed in Fig. 8(a). Instead, it chooses to accelerate to 40 km/h while remaining in the current lane. This decision reflects the system's assessment that although the lead vehicle is moving at a low speed, 30km/h, the gap is sufficiently large, making it safe to accelerate while still addressing the occupant's desire for a more assertive driving response. The underlying reasoning and decision-making process of PACE-ADS in this scenario is detailed in Fig. 8(g).

Each test run was repeated 30 times with different random seeds, which introduced variability in the speed profile of the leading vehicle. This randomness affected both the timing and magnitude of its acceleration and deceleration. The vehicle's initial speed was set to 40 km/h, while its maximum and minimum speeds were limited to 60 km/h and 30 km/h, respectively. The average results are summarized in Table 4, including average speed ($\bar{v}$) over the entire car-following process, the average absolute acceleration ($|\ \bar{a}\ |$) during the acceleration phase, and the average absolute deceleration ($|\ \bar{b}\ |$) during the deceleration phase. The ego vehicle's average speed decreases progressively from very impatient (54.8 m/s) to very anxious (30.2 m/s), while the acceleration during the speeding phase becomes increasingly smoother (from $1.39\ m/s^2$ to $0.38\ m/s^2$). For the deceleration phase, there are no values for the very impatient and impatient states because the PACE-ADS, after accelerating to a higher speed, chose to change lanes and cruise instead of slowing down. In contrast, for the very anxious condition, the average deceleration ($1.39\ m/s^2$) is greater than that in the normal and anxious states, as PACE-ADS determines that reducing speed quickly helps to alleviate the passenger's anxiety, which is consistent with the results illustrated in Fig. 8(f).

Table 4. Personalized evaluation results in the car following scenario (averaged over 30 runs).

| Occupant state | Metrics | | |
|---|---|---|---|
| | $\bar{v}(m/s)$ | $\|\ \bar{a}\ \|\ (m/s^2)$ | $\|\ \bar{b}\ \|\ (m/s^2)$ |
| Very impatient | 54.8 | 1.39 | \ |
| Impatient | 46.1 | 0.91 | \ |
| Normal | 42.5 | 0.56 | 0.98 |
| Anxious | 34.4 | 0.39 | 0.43 |
| Very anxious | 30.2 | 0.38 | 1.39 |

Note: The leading vehicle's average speed, acceleration, and deceleration are $45.7\ m/s$, $0.72\ m/s^2$, and $1.17\ m/s^2$ respectively.

**D. Immobilization Recovery**

In this section, we design three additional scenarios to evaluate how PACE-ADS recovers an immobilized AV by selecting appropriate driving behaviors based on occupant instructions, thereby reducing travel delays and improving traffic efficiency. Although the Bench2Drive benchmark [31] covers a wide range of driving tasks, it does not include sufficient edge cases that challenge recovery and decision-making capabilities. The three scenarios introduced here are not meant to be comprehensive but are representative cases that allow assessment of how PACE-ADS interprets ambiguous environments, makes decisions under uncertainty, and interacts with the occupant when necessary.

**Static obstacles scenario.** The ego AV cruises along a street in CARLA's city map at a speed of 20 km/h, with several vehicles parked on the right shoulder. Two paper bags are detected in the AV's lane about 5 meters ahead. Upon detection, the AV initiates a stop behavior to avoid a potential collision, as shown by the green curve without markers in Figure 9. After stopping, the coordinator senses that the AV has halted unexpectedly, even though the destination has not been reached and no nearby traffic control devices require stopping. The coordinator activates the driver agent to analyze the surrounding traffic environment. The driver agent correctly identifies the objects and reports this information to the coordinator. However, due to insufficient information about the contents of these bags, the coordinator cannot determine whether to drive over or not. To prioritize safety, it conservatively treats the bags as a potential hazard and halts the AV. The coordinator then communicates the reason for stopping to the passenger and seeks guidance. After the passenger confirms that the bags are empty and pose no danger, PACE-ADS overrides the initial stop behavior and resumes operation, as shown by the green curve with markers in Figure 9. This scenario illustrates PACE-ADS's ability to detect and analyze immobilized AV behavior and to proactively involve the passenger when uncertainty arises.

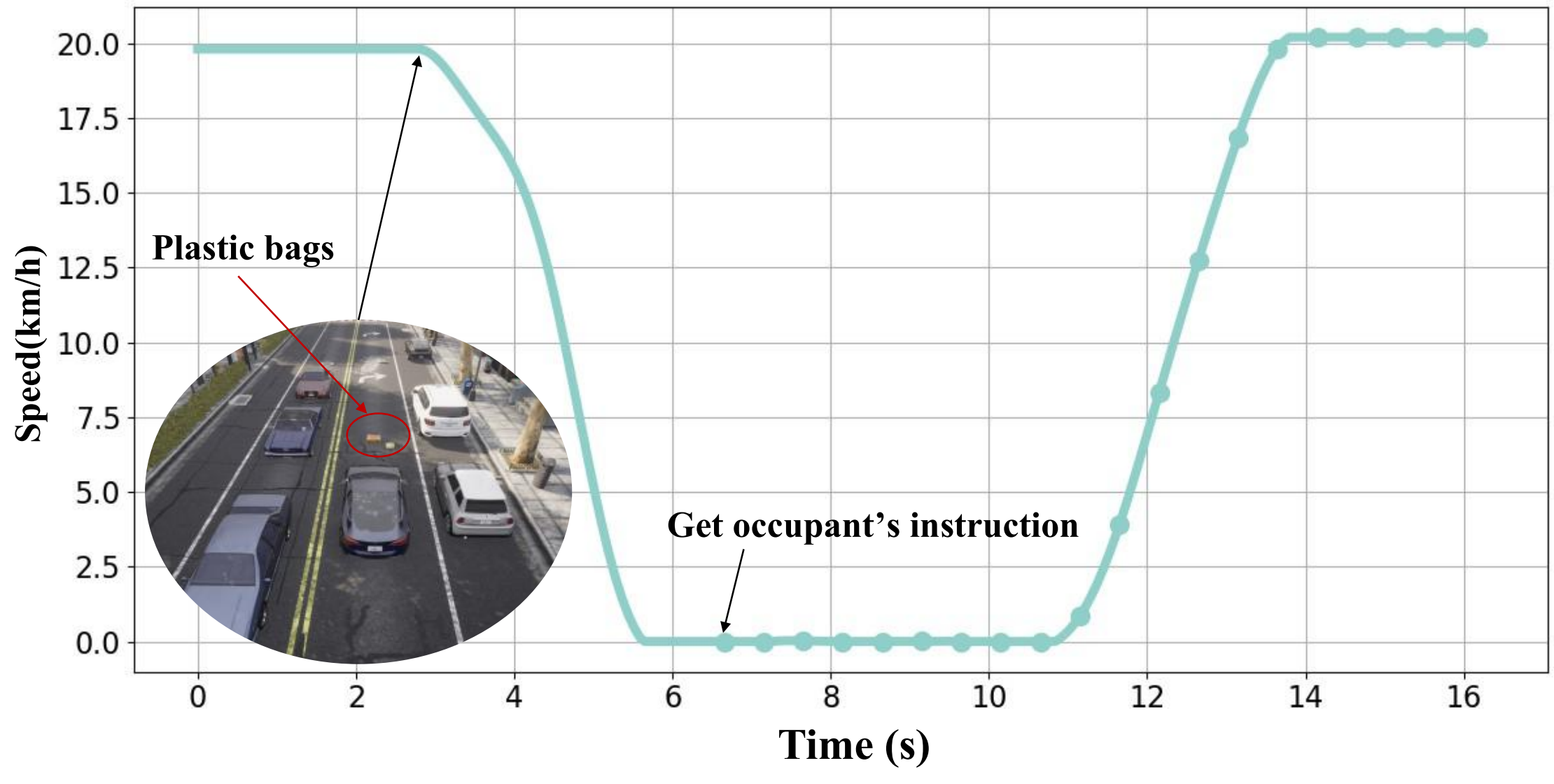

(a) AV's speed profile.

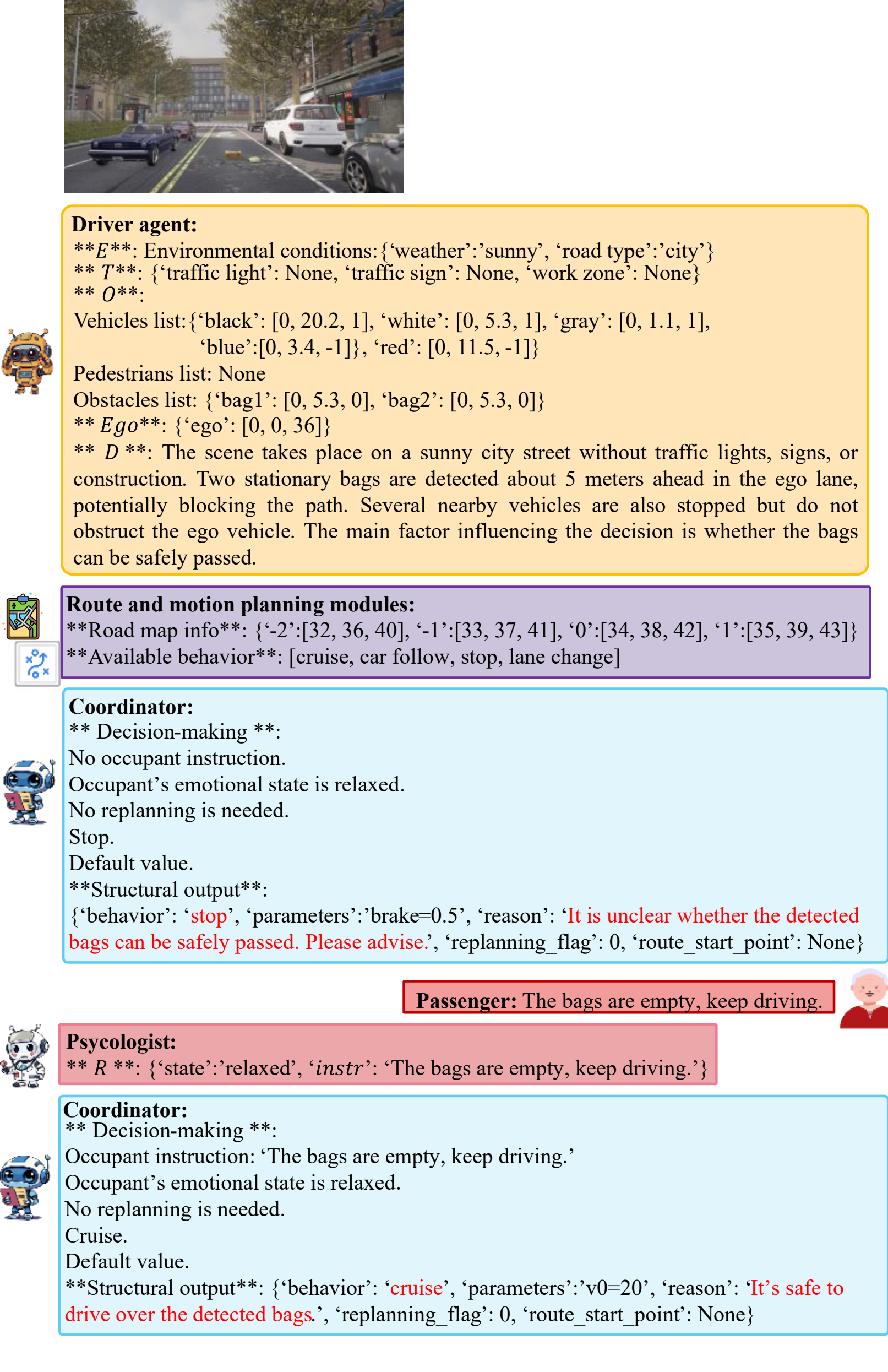

(b) PACE-ADS reasoning outputs.

Figure 9. Static obstacles experiment results.

**Roundabout scenario.** The AV system's route planning module is forced to generate a path that continuously loops a roundabout, simulating an immobilization case similar to the Waymo Robotaxi incident. This behavior is illustrated by the green curve without markers in Figure 10.

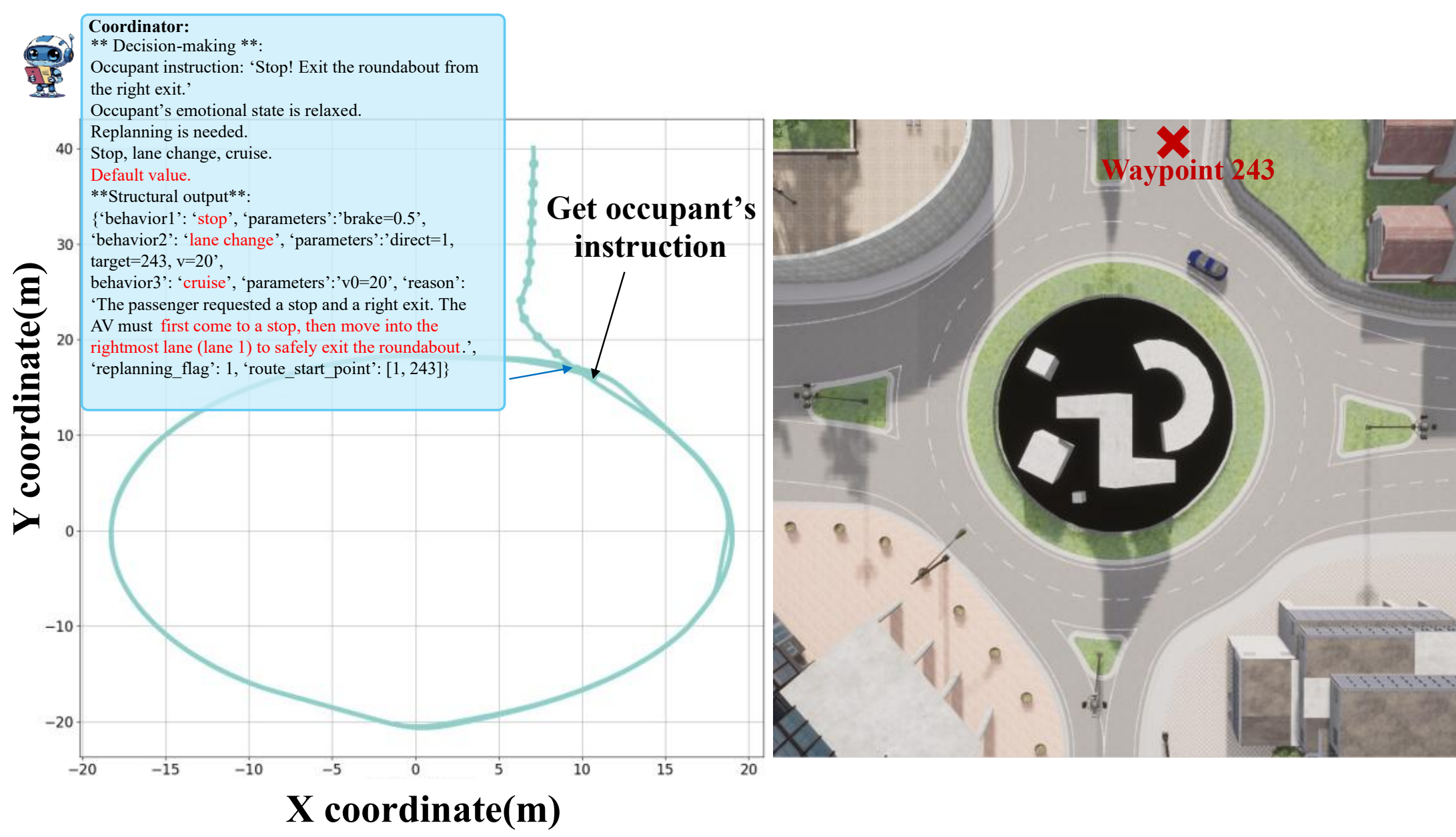

Figure 10. Roundabout experiment results.

After the vehicle completes its fifth loop, passenger guidance is issued to PACE-ADS: "Stop! Exit the roundabout from the right exit." Upon receiving this instruction, the coordinator re-analyzes the situation. It parses the guidance into actionable components, including stop, lane change, and cruise, and performs route replanning. A new route starting point is marked by the red cross in Figure 10. Following this decision, the AV successfully exits the roundabout, shown by the green curve with markers in Figure 10. This scenario demonstrates the capability of PACE-ADS to interpret passenger commands. It extracts the intended meaning, decomposes the instruction into semantic actions, matches them with available driving behaviors, and generates a coherent sequence of decision-making steps to assist the AV in resuming normal operations.

**Road closure scenario.** Next, we design a road closure scenario in which the AV travels from the origin (black triangle) to the destination (red cross), as shown in Figure 11. While following the planned route (black dashed curve), the AV detects obstacles ahead and decides to stop. After stopping, the coordinator detects that the AV has stopped unexpectedly, despite not having reached its destination and with no nearby traffic signals or stop signs. In response, the coordinator activates the driver agent to analyze the surrounding traffic environment.

The driver agent detects that barricades fully block all travel lanes. Based on this input, the coordinator decides to maintain the stopping state and prompts the passenger to confirm whether a new route should be planned with an explanation. Upon receiving passenger confirmation, the coordinator queries the map data from the route planning module and selects waypoint 430 in the left lane (marked by the red triangle) as the new starting point for replanning. The newly generated route to the destination is shown as a red solid line in Figure 11. This scenario illustrates PACE-ADS's capabilities in perceiving and interpreting the driving environment, providing explanations for vehicle behavior, and proactively engaging the passenger in decision-making.

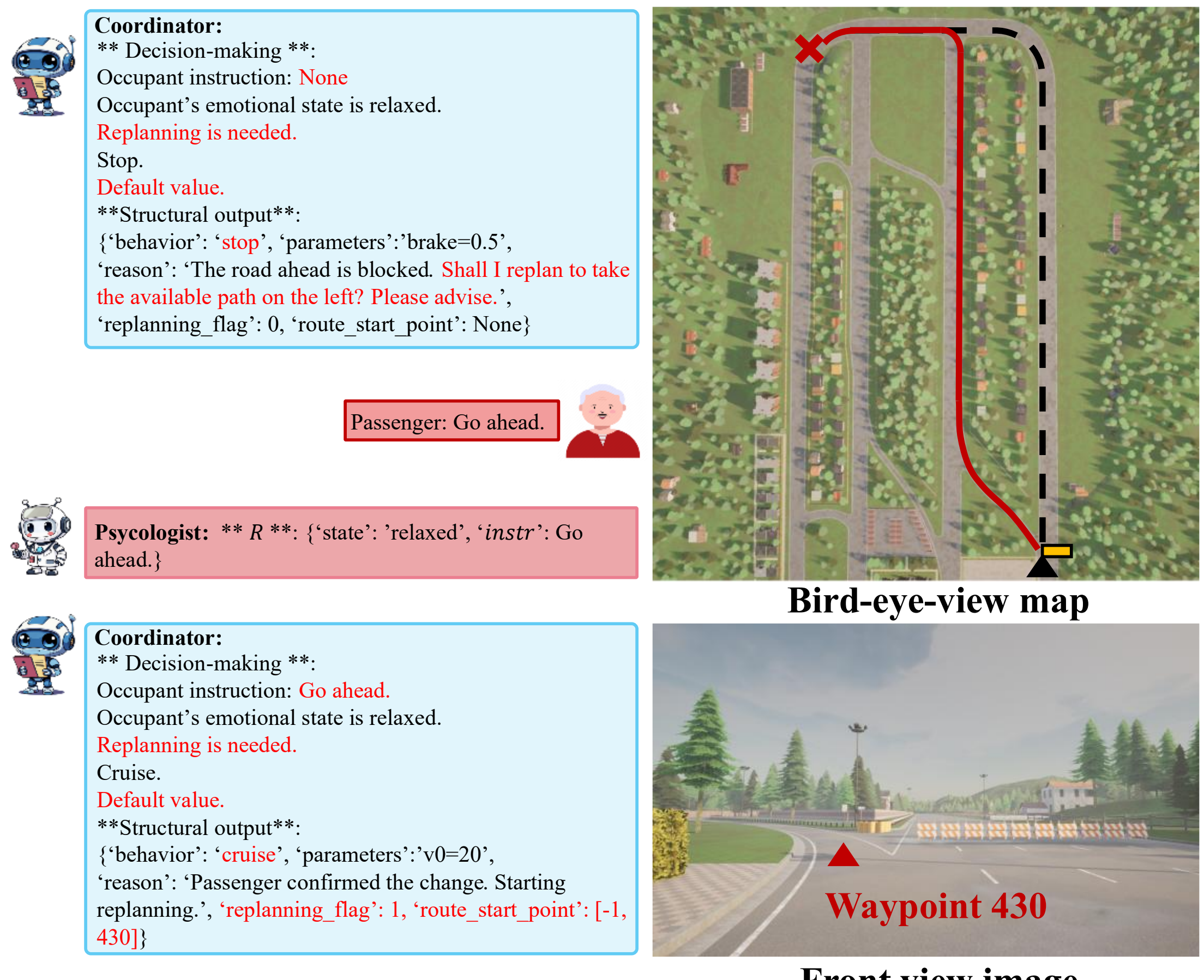

Figure 11. Road closure experiment results.

## IV. Conclusion

This paper proposes PACE-ADS, a novel framework for human-centered autonomy that integrates LLMs to support adaptive and interpretable driving behavior. Simulation results demonstrate that PACE-ADS effectively adapts to diverse emotional states and situational contexts, adjusting driving behavior to align with occupant preferences while maintaining safety. Furthermore, in scenarios where the AV becomes immobilized, PACE-ADS is capable of resolving the situation by following natural language guidance provided by the occupant. PACE-ADS redefines the occupant's role from passive observer to active participant. By detecting emotional states and engaging the occupant during ambiguous situations, it reduces discomfort, confusion, and risk in non-standard driving scenarios where rigid automation often falls short. Its interpretability through natural language dialogue and responsiveness to individual preferences enhances user trust, perceived control, and acceptance of AVs. Moreover, PACE-ADS is modular and compatible with existing AV stacks, enabling seamless integration without major architectural changes. This makes it a scalable enhancement for next-generation AVs.

Future research will focus on deploying the PACE-ADS framework on physical AV platforms to evaluate its effectiveness under real-world conditions. Evaluation will involve real-time monitoring of psychological states alongside post-drive interviews to assess perceived personalization, comfort, and trust. Lightweight LLM architectures optimized for edge deployment will be explored. Reducing computational demands is essential for operation on embedded AV systems. Techniques such as model distillation, pruning, and quantization will be employed to enable low-latency inference while preserving the interpretability and responsiveness required for human-centered autonomy. To enhance long-term adaptability, future development will also incorporate a memory module into PACE-ADS, enabling the system to learn from user interactions over time. Such memory-enabled architecture will allow the LLM to retain contextual information across sessions, improving consistency in decision-making and supporting user-specific behavioral adaptation, particularly important in private vehicle applications.

**Reference**

[1] Strategic Market Research. (2022). *Advanced Driver Assistance Systems (ADAS) Market: By System Type, Component Type, Vehicle Type, Electric Vehicle, Region – Global Growth Forecast 2021-2030*. Retrieved from https://www.strategicmarketresearch.com/market-report/adas-market

[2] Schrum, M. L., Sumner, E., Gombolay, M. C., & Best, A. (2024). Maveric: A data-driven approach to personalized autonomous driving. *IEEE Transactions on Robotics*, *40*, 1952-1965.

[3] Su, C., Deng, W., He, R., Wu, J., & Jiang, Y. (2018). *Personalized adaptive cruise control considering drivers' characteristics* (No. 2018-01-0591). SAE Technical Paper.

[4] Han, Z., Zheng, X., Ren, Y., Li, X., & Wang, Q. (2023). Personalized Adaptive Cruise Control with Deep Reinforcement Learning. *Ergonomics In Design*, *77*(77).

[5] Zhu, B., Yan, S., Zhao, J., & Deng, W. (2018). Personalized lane-change assistance system with driver behavior identification. *IEEE Transactions on Vehicular Technology*, *67*(11), 10293-10306.

[6] Wang, Y., Wang, Z., Han, K., Tiwari, P., & Work, D. B. (2022). Gaussian process-based personalized adaptive cruise control. *IEEE Transactions on Intelligent Transportation Systems*, *23*(11), 21178-21189.

[7] Nava, D., Panzani, G., Zampieri, P., & Savaresi, S. M. (2019, October). A personalized Adaptive Cruise Control driving style characterization based on a learning approach. In *2019 IEEE Intelligent Transportation Systems Conference (ITSC)* (pp. 2901-2906). IEEE.

[8] Gao, B., Cai, K., Qu, T., Hu, Y., & Chen, H. (2020). Personalized adaptive cruise control based on online driving style recognition technology and model predictive control. *IEEE transactions on vehicular technology*, *69*(11), 12482-12496.

[9] Lee, H., & Samuel, S. (2024). Classification of User Preference for Self-Driving Mode and Behaviors of Autonomous Vehicle. *IEEE Transactions on Intelligent Vehicles*.
[10] Ling, J., Li, J., Tei, K., & Honiden, S. (2021, December). Towards personalized autonomous driving: An emotion preference style adaptation framework. In *2021 IEEE International Conference on Agents (ICA)* (pp. 47-52). IEEE.
[11] Lillicrap, T. P., Hunt, J. J., Pritzel, A., Heess, N., Erez, T., Tassa, Y., ... & Wierstra, D. (2015). Continuous control with deep reinforcement learning. *arXiv preprint arXiv:1509.02971*.
[12] Bao, N., Capito, L., Yang, D., Carballo, A., Miyajima, C., & Takeda, K. (2022). Data-driven risk-sensitive control for personalized lane change maneuvers. *IEEE Access*, *10*, 36397-36415.
[13] Chen, X., Zhai, Y., Lu, C., Gong, J., & Wang, G. (2017, June). A learning model for personalized adaptive cruise control. In *2017 IEEE intelligent vehicles symposium (IV)* (pp. 379-384). IEEE.
[14] Li, D., & Liu, A. (2023). Personalized lane change decision algorithm using deep reinforcement learning approach. *Applied Intelligence*, *53*(11), 13192-13205.
[15] Han, Z., Zheng, X., Ren, Y., Li, X., & Wang, Q. (2023). Personalized Adaptive Cruise Control with Deep Reinforcement Learning. *Ergonomics In Design*, *77*(77).
[16] Zhao, Z., Liao, X., Abdelraouf, A., Han, K., Gupta, R., Barth, M. J., & Wu, G. (2023, October). Real-time learning of driving gap preference for personalized adaptive cruise control. In *2023 IEEE International Conference on Systems, Man, and Cybernetics (SMC)* (pp. 4675-4682). IEEE.
[17] Tian, H., Wei, C., Jiang, C., Li, Z., & Hu, J. (2022). Personalized lane change planning and control by imitation learning from drivers. *IEEE Transactions on Industrial Electronics*, *70*(4), 3995-4006.
[18] Doudou, M., Bouabdallah, A., & Berge-Cherfaoui, V. (2020). Driver drowsiness measurement technologies: Current research, market solutions, and challenges. International Journal of Intelligent Transportation Systems Research, 18, 297-319.
[19] Zhang, Z., Ning, H., & Zhou, F. (2022). A systematic survey of driving fatigue monitoring. *IEEE transactions on intelligent transportation systems*, *23*(11), 19999-20020.
[20] Kashevnik, A., Shchedrin, R., Kaiser, C., & Stocker, A. (2021). Driver distraction detection methods: A literature review and framework. *IEEE Access*, *9*, 60063-60076.
[21] Misra, A., Samuel, S., Cao, S., & Shariatmadari, K. (2023). Detection of driver cognitive distraction using machine learning methods. *IEEE Access*, *11*, 18000-18012.
[22] Baldini, F., Tariq, F. M., Bae, S., & Isele, D. (2024). Don't Get Stuck: A Deadlock Recovery Approach. *arXiv preprint arXiv:2408.10167*.
[23] Honda, K., Yonetani, R., Nishimura, M., & Kozuno, T. (2024, May). When to replan? an adaptive replanning strategy for autonomous navigation using deep reinforcement learning. In *2024 IEEE International Conference on Robotics and Automation (ICRA)* (pp. 6650-6656). IEEE.
[24] Schumann, O., Buchholz, M., & Dietmayer, K. (2023, September). Efficient Path Planning in Large Unknown Environments with Switchable System Models for

Automated Vehicles. In *2023 IEEE 26th International Conference on Intelligent Transportation Systems (ITSC)* (pp. 2466-2472). IEEE.
[25] Schmittle, M., Baijal, R., Hou, B., Srinivasa, S., & Boots, B. (2024, May). Multi-Sample Long Range Path Planning under Sensing Uncertainty for Off-Road Autonomous Driving. In *2024 IEEE International Conference on Robotics and* Automation (ICRA) (pp. 4598-4604). IEEE.
[26] King, R.E. (2024). Empty Waymo robotaxi loops roundabout 37 times as if it was stuck in logic loop. *Jalopnik*. Retrieved from https://jalopnik.com/empty-waymo-robotaxi-loops-roundabout-37-times-as-if-it-1851721933
[27] NTDTV. (2024). "萝卜快跑"跑不了致交通拥堵 武汉交警束手无策（视频）. 新唐人电视台. Retrieved from https://www.ntdtv.com/gb/2024/07/12/a103896787.html
[28] Kondyli, V., Bhatt, M., Levin, D., & Suchan, J. (2023). How do drivers mitigate the effects of naturalistic visual complexity? On attentional strategies and their implications under a change blindness protocol. Cognitive research: principles and implications, 8(1), 54.
[29] M. Harris, "Tesla's Autopilot Depends on a Deluge of Data," IEEE Spectrum, Aug. 4, 2022. [Online]. Available: https://spectrum.ieee.org/tesla-autopilot-data-deluge
[30] Bao, Z., & Li, Q. (2025). Large Language Model-assisted Autonomous Vehicle Recovery from Immobilization. *arXiv preprint.*
[31] Jia, X., Yang, Z., Li, Q., Zhang, Z., & Yan, J. (2024). Bench2drive: Towards multi-ability benchmarking of closed-loop end-to-end autonomous driving. *Advances in Neural Information Processing Systems*, *37*, 819-844.
[32] Lüddecke, T., & Ecker, A. (2022). Image segmentation using text and image prompts. In Proceedings of the IEEE/CVF conference on computer vision and pattern recognition (pp. 7086-7096).
[33] Hu, A., Russell, L., Yeo, H., Murez, Z., Fedoseev, G., Kendall, A., ... & Corrado, G. (2023). Gaia-1: A generative world model for autonomous driving. arXiv preprint arXiv:2309.17080.
[34] Xu, Z., Zhang, Y., Xie, E., Zhao, Z., Guo, Y., Wong, K. Y. K., ... & Zhao, H. (2024). Drivegpt4: Interpretable end-to-end autonomous driving via large language model. *IEEE Robotics and Automation Letters*.
[35] Fang, S., Liu, J., Ding, M., Cui, Y., Lv, C., Hang, P., & Sun, J. (2025). Towards interactive and learnable cooperative driving automation: a large language model-driven decision-making framework. *IEEE Transactions on Vehicular Technology*.
[36] Wang, Y., Jiao, R., Zhan, S. S., Lang, C., Huang, C., Wang, Z., ... & Zhu, Q. (2023). Empowering autonomous driving with large language models: A safety perspective. arXiv preprint arXiv:2312.00812.
[37] Cui, C., Ma, Y., Cao, X., Ye, W., & Wang, Z. (2024). Drive as you speak: Enabling human-like interaction with large language models in autonomous vehicles. In *Proceedings of the IEEE/CVF Winter Conference on Applications of Computer Vision* (pp. 902-909).
[38] Mao, J., Qian, Y., Ye, J., Zhao, H., & Wang, Y. (2023). Gpt-driver: Learning to drive with gpt. *arXiv preprint arXiv:2310.01415*.

[39] Sha, H., Mu, Y., Jiang, Y., Chen, L., Xu, C., Luo, P., ... & Ding, M. Languagempc: Large language models as decision makers for autonomous driving. arXiv 2023. *arXiv preprint arXiv:2310.03026*.
[40] Wu, D., Han, W., Liu, Y., Wang, T., Xu, C. Z., Zhang, X., & Shen, J. (2025, April). Language prompt for autonomous driving. In *Proceedings of the AAAI Conference on Artificial Intelligence* (Vol. 39, No. 8, pp. 8359-8367).
[41] Wen, L., Fu, D., Li, X., Cai, X., Ma, T., Cai, P., ... & Qiao, Y. (2023). Dilu: A knowledge-driven approach to autonomous driving with large language models. *arXiv preprint arXiv:2309.16292*.
[42] Cui, C., Yang, Z., Zhou, Y., Peng, J., Park, S. Y., Zhang, C., ... & Wang, Z. (2024). On-Board Vision-Language Models for Personalized Autonomous Vehicle Motion Control: System Design and Real-World Validation. *arXiv preprint arXiv:2411.11913*.
[43] Wei J, Wang X, Schuurmans D, et al. Chain-of-thought prompting elicits reasoning in large language models[J]. Advances in neural information processing systems, 2022, 35: 24824-24837.
[44] Hao, X., Diao, Y., Wei, M., Yang, Y., Hao, P., Yin, R., ... & Liu, Y. (2025). MapFusion: A novel BEV feature fusion network for multi-modal map construction. *Information Fusion*, *119*, 103018.
[45] Jamali, M., Davidsson, P., Khoshkangini, R., Ljungqvist, M. G., & Mihailescu, R. C. (2025). Context in object detection: a systematic literature review. *Artificial Intelligence Review*, *58*(6), 1-89.
[46] Hurst A, Lerer A, Goucher A P, et al. Gpt-4o system card[J]. arXiv preprint arXiv:2410.21276, 2024.
[47] Macenski S, Foote T, Gerkey B, et al. Robot operating system 2: Design, architecture, and uses in the wild[J]. Science robotics, 2022, 7(66): eabm6074.
[48] Dosovitskiy A, Ros G, Codevilla F, et al. CARLA: An open urban driving simulator[C]//Conference on robot learning. PMLR, 2017: 1-16.
[49] Hart, P. E., Nilsson, N. J., & Raphael, B. (1972). Correction to" a formal basis for the heuristic determination of minimum cost paths". *ACM SIGART Bulletin*, (37), 28-29.
[50] Kesting A, Treiber M, Helbing D. Enhanced intelligent driver model to access the impact of driving strategies on traffic capacity[J]. Philosophical Transactions of the Royal Society A: Mathematical, Physical and Engineering Sciences, 2010, 368(1928): 4585-4605.
[51] Saganowski S, Komoszyńska J, Behnke M, et al. Emognition dataset: emotion recognition with self-reports, facial expressions, and physiology using wearables[J]. Scientific data, 2022, 9(1): 158.
[52] Radford, A., Kim, J. W., Xu, T., Brockman, G., McLeavey, C., & Sutskever, I. (2023, July). Robust speech recognition via large-scale weak supervision. In *International conference on machine learning* (pp. 28492-28518). PMLR.

## Appendix

Table 1. Description of the input and output data for the driver agent, psychologist, and coordinator.

| Module | Input/ Output | Data | Example | Description |
| --- | --- | --- | --- | --- |
| Driver agent | Input | $I_{front}$, $I_{bev}$ | RGB images | Captured by front view camera and BEV camera. |
| | | $\boldsymbol{O^p} = \{\ldots O_i^p = \{id, dist, v, loc\} \ldots\}$ | {…, $O_i^p$= {'black vehicle', 20m, 35km/h, [1, 243]}, …} | Analytics from the vehicle's perception module. The object measurement data $\boldsymbol{O^p}$ includes information for each surrounding object $i$: its class or identifier, relative distance, velocity, and spatial location (current lane id and waypoint id). |
| | Output | $E = \{$'weather': $\ldots,$ 'road type': $\ldots\}$ | {'weather':'sunny', 'road type':'city'} | Weather and road type affect driving speed. |
| | | $T = \{$'traffic light': $\ldots,$ 'traffic sign': $\ldots,$ 'work zone': …$\}$ | {'traffic light':[red, (10,20)], 'traffic sign': 'stop', 'work zone': [(30,10), affect?]} | Traffic light's state and traffic sign's content are observed from image. The coordinates of traffic lights and work zone are obtained from perception module. The affect? represents whether the work zone affects AV's driving. |
| | | $\boldsymbol{O} = \{$Vehicles list: $\{$ 'id': [speed, distance, lane id], $\ldots\}$, Pedestrians list: $\{$ 'id': [speed, distance, lane id], $\ldots\}$, Obstacles list: $\{$ 'id': [speed, distance, lane id], $\ldots\}\}$ | Vehicles list: {'black': [35km/h, 20m, 0], …},<br>Pedestrians list: {'red': [4km/h, 10m, 1], …}<br>Obstacles list: {'barrier': [0km/h, 6m, -1], …} | 'black' represents the vehicle's color, which is also the vehicle's role name offered by perception module. The speed, distance, lane id are obtained from perception module.<br>The format of data in pedestrians list is the same as vehicles list.<br>'barrier' is the role name of the obstacle offered by perception module. |
| | | $Ego = \{v, loc\}$ | {45km/h, [0, 243]} | AV's velocity, spatial location (current lane id and waypoint id). |
| | | $D$ | The scene is on a sunny city street with no traffic lights, signs, or construction. Two stationary bags are detected about 5.3 meters ahead in the ego lane, possibly blocking the path. Several nearby vehicles are also stationary, but they do not directly interfere with the ego vehicle. | A scene description emphasizing traffic semantics, focusing on how environmental conditions and traffic control elements affect driving, and describing traffic participants and obstacles, noting their interaction with the ego vehicle and potential impact. |

| Psychologist | Input | $EEG = \{ Alpha_{AF7}: [\ldots], Alpha_{AF8}: [\ldots], Beta_{AF7}: [\ldots], Beta_{AF8}: [\ldots], Theta_{AF7}: [\ldots], Theta_{AF8}: [\ldots]\}$ | {[0.0864, …], [0.6023, …], [0.3069, …], [0.4573, …], [0.1741, …], [0.5625, …]} | EEG signals and heart rate are extracted from the Emognition dataset. They are time-synchronized and collectively respond to the same type of interest. |
|---|---|---|---|---|
| | | $HR$ | [68, 68, 69, 69, 69, 70, …] | |
| | | $I_{face}$ | RGB image | Images are extracted from videos offered by Emognition dataset, and together with EEG and heart rate signals, reflect the same emotional state. |
| | | $Ver$ | 'The bags are empty, keep driving.' | Occupant's verbal commands. |
| | Output | $R = \{state, instr\}$ | {'state': 'relaxed', 'instr':'None'} | The $state$ is the passenger's emotions, such as impatience, relaxation, or very anxiety. $instr$ is occupant's instructions. |
| Coordinator | Input | $\boldsymbol{Y} = \{E, T, \boldsymbol{O}, Ego, D\}$ | {E = {…}, T = {…}, **O** = {…}, Ego = {…}, D = {…}} | Structural output from driver agent. |
| | | $R = \{state, instr\}$ | {'state': 'relaxed', 'instr':'None'} | Structural output from psychologist. |
| | | $M$ = {'left lane id': [waypoint id], 'current lane id': [waypoint id], 'right lane id': …} | {'-1': [17, 18, 19], '0': [21, 22, 23], '1': [25, 26, 27]} | The AV's current lane id is 0, with negative values representing left lanes and positive values representing right lanes. |
| | | $\boldsymbol{A_{lib}}$ = {{ 'behavior1's introduction': …, 'parameters description': …, 'default value': … }, …} | {'behavior1's introduction': …, 'parameters description': There are four parameters…, 'default value': v0=40, …} | $\boldsymbol{A_{lib}}$ includes an introduction to the behavior's functions, the parameters associated with the algorithm implementing the behavior, and the corresponding default parameter values. |
| | Output | $N$ = {'behavior': …, 'parameters': …, 'reason': …, 'replanning flag': …, 'route start point': [lane id, wp id]} | {'behavior': 'car following', 'parameters': 'v0=40, t=1.5, a=0.8, b=1.67, s0=3', 'reason': '…', 'replanning_flag': 0, 'route_start_point': [0, 243]} | The structural output includes the selected behavior, parameter values, and the reasoning behind the decision. The replanning_flag is set to 0 or 1, indicating whether replanning is needed. lane id and wp id represent the lane and waypoint where the replanning starts. |